\documentclass[journal,twoside,web]{ieeecolor}
\usepackage{generic}
\usepackage{cite}
\usepackage{amsmath,amssymb,amsfonts}
\usepackage{algorithmic}
\usepackage{graphicx}
\usepackage{epstopdf}   
\usepackage{textcomp}
\usepackage{multirow}
\usepackage{booktabs}
\usepackage{color}
\usepackage{bm}
\usepackage{subfigure}
\usepackage{xcolor}
\usepackage{soul}
\definecolor{my_yellow}{RGB}{255,255,0}
\sethlcolor{my_yellow}
\soulregister{\cite}7
\soulregister{\citep}7
\soulregister{\citet}7
\soulregister{\ref}7
\soulregister{\pageref}7

\def\BibTeX{{\rm B\kern-.05em{\sc i\kern-.025em b}\kern-.08em
    T\kern-.1667em\lower.7ex\hbox{E}\kern-.125emX}}
\begin{document}
\title{Learn Fine-grained Adaptive Loss for Multiple Anatomical Landmark Detection in Medical Images}
\author{Guang-Quan Zhou\textsuperscript{$\star$}, \IEEEmembership{Member, IEEE}, Juzheng Miao, Xin Yang\textsuperscript{$\star$}, Rui Li, En-Ze Huo, Wenlong Shi, Yuhao Huang, Jikuan Qian, Chaoyu Chen, Dong Ni\textsuperscript{$\star$}
\thanks{This work was supported by the National Natural Science Foundation of China (NSFC61771130), and the National Key R\&D Program of China (2018YFA0704102). Asterisk indicates corresponding author.}
\thanks{Guang-Quan Zhou, Juzheng Miao, and En-Ze Huo are with the State Key Laboratory of Bioelectronics, School of Biological Science and Medical Engineering, Southeast University, Nanjing, China (email: guangquan.zhou@seu.edu.cn).}
\thanks{Xin Yang, Rui Li, Wenlong Shi, Yuhao Huang, Jikuan Qian, Chaoyu Chen, and Dong Ni are with the National-Regional Key Technology Engineering Laboratory for Medical Ultrasound, Guangdong Key Laboratory for Biomedical Measurements and Ultrasound Imaging, School of Biomedical Engineering, Health Science Center, Shenzhen University, Shenzhen, China, and also with the Medical UltraSound Image Computing (MUSIC) Lab, Shenzhen, China (email:yangxinknow@gmail.com, nidong@szu.edu.cn).}}

\maketitle

\begin{abstract}
Automatic and accurate detection of anatomical landmarks is an essential operation in medical image analysis with a multitude of applications. Recent deep learning methods have improved results by directly encoding the appearance of the captured anatomy with the likelihood maps (i.e., heatmaps). However, most current solutions overlook another essence of heatmap regression, the objective metric for regressing target heatmaps and rely on hand-crafted heuristics to set the target precision, thus being usually cumbersome and task-specific. In this paper, we propose a novel learning-to-learn framework for landmark detection to optimize the neural network and the target precision simultaneously. The pivot of this work is to leverage the reinforcement learning (RL) framework to search objective metrics for regressing multiple heatmaps dynamically during the training process, thus avoiding setting problem-specific target precision. We also introduce an early-stop strategy for active termination of the RL agent's interaction that adapts the optimal precision for separate targets considering exploration-exploitation tradeoffs. This approach shows better stability in training and improved localization accuracy in inference. Extensive experimental results on two different applications of landmark localization: 1) our in-house prenatal ultrasound (US) dataset and 2) the publicly available dataset of cephalometric X-Ray landmark detection, demonstrate the effectiveness of our proposed method. Our proposed framework is general and shows the potential to improve the efficiency of anatomical landmark detection.
\end{abstract}


\footnote{This article has been accepted for publication in a future issue of this journal, but has not been fully edited. Content may change prior to final publication. Citation information: DOI 10.1109/JBHI.2021.3080703, IEEE Journal of Biomedical and Health Informatics.}

\section{Introduction}
\label{sec:introduction}
\begin{figure}[!t]
 \centerline{\includegraphics[width=\columnwidth]{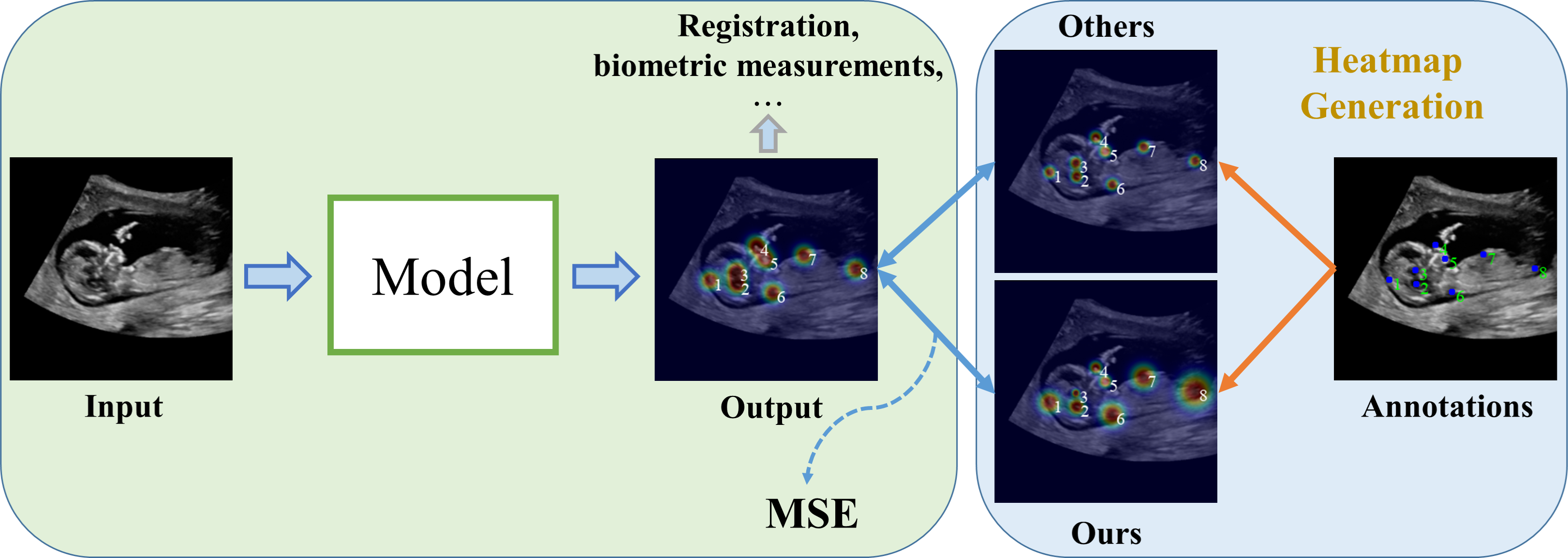}}
 \caption{Illustration of differences between our method and current heatmap regression methods. Current methods tend to use fixed Gaussian distributions for ground truth heatmap, while adaptive standard deviations of Gaussian distributions are utilized in our proposed method.}
 \label{intro}
\end{figure}
Accurate anatomical landmark detection is essential in many medical image applications, e.g., for registration or biometric measurements. The data-driven deep neural networks (DNNs) have revolutionized many medical image analysis tasks by effectively exploiting large annotated medical image databases \cite{shen2017deep, khan2020survey, huang2020segmentation, yin2019domain}. Anatomical landmark localization has also seen tremendous progress in recent years due to the widespread availability of deep learning models. The advent of these automated solutions brings opportunities for landmark localization in medical images to achieve fast and accurate diagnostic biometric measurements \cite{2016A, ghesu2017multi, du2018articulated, chen2019cephalometric, zhou2020single, zhang2017alzheimer, guo2020robust}, obviating the time-consuming and subjective manual identification.

Most current solutions formulate the problem as a structured output regression problem to localize the landmark directly from the images \cite{wu2017facial, newell2016stacked}. Some of them directly regress the image coordinates where the map from the input appearance to the image coordinates is learned. However, this complicated mapping makes the direct regression of coordinates usually difficult to train \cite{teixeira2019adaloss, noothout2020deep, pfister2015flowing}. Moreover, a single target output for each landmark location instead of several candidate locations might cause missing landmarks. As a result, regressing landmarks directly is less usable as a pre-processing step of more complex vision tasks \cite{teixeira2019adaloss}. Some reinforcement learning (RL) methods directly explore the landmark location by learning to generate navigation trajectories that point towards the sought-for landmarks with a multi-scale strategy \cite{ghesu2017multi, alansary2019evaluating} or a dedicated reward design \cite{zhang2020enhanced}. Nonetheless, these RL-based methods require huge datasets and computationally extensive training. Recent works tend to regress the likelihood maps (i.e., heatmaps) to encode semantics and spatial context information for inferring landmark locations, showing effectiveness in multiple domains \cite{chen2019cephalometric, newell2016stacked, xiao2018simple, arik2017fully, chen2018cascaded, sun2019deep, payer2016regressing, payer2019integrating}. For instance, Newell et al. \cite{newell2016stacked} employ a stacked encoder-decoder convolutional neural network (CNN), called the Hourglass network to regress heatmaps at the pixel level for landmark inference by consolidating features across multiple scales. These DNN models usually exploit high-resolution feature maps for the heatmap regression to improve landmark localization performance. Payer et al. \cite{payer2016regressing, payer2019integrating} combine the local appearance with the global spatial configuration for multiple landmark localization in medical images. A set of parallel or cascaded multi-scale subnetworks is also connected to fuse multi-scale representations, leading to high-resolution feature maps \cite{chen2018cascaded, sun2019deep}. Similarly, Chen et al. \cite{chen2019cephalometric} incorporate the self-attention mechanism into pyramid architecture to get a high-resolution and semantically enhanced fusion feature for the heatmap regression. On the other hand, the heatmaps ground truth of each landmark is another essence since it determines the accuracy of estimating pixel intensity and could cause inaccurate landmark localization. However, previous works rarely studied this objective metric against the heatmap of each landmark.

As is shown in Fig. \ref{intro}, current heatmap regression methods employ the multivariate Gaussian distributions centered at each landmark location as the ground truth of the heatmaps in training. These models widely use the Mean Square Error (MSE) loss to estimate pixel intensity at the mode of the Gaussian as the target precision to predict the landmark localization. Thus, the objective metric for evaluating the ground truth heatmap based on Gaussian distributions during training plays a vital role in locating landmarks, significantly influencing landmark detection accuracy and convergence. Nevertheless, there is a tradeoff to appropriately set the variance of Gaussian distributions to define these target heatmaps in training \cite{payer2019integrating, teixeira2019adaloss}. Firstly, there is an inherited uncertainty and inconsistency among landmarks during the training \cite{lampert2016empirical}. However, all target landmarks usually set the same variance or standard deviation of Gaussian during training in previous literature. Moreover, a small standard deviation in Gaussian mode could contribute to localizing landmarks accurately, but it is challenging to train the network due to the highly sparse output distributions. By contrast, a large variance is more comfortable to train with the cost of prediction errors, resulting in a blurry and dilated predicted heatmap with low intensity. Recently, an application-independent Adaloss indirectly manipulates landmark localization objectives by gradually decreasing the variance for each target separately throughout the training \cite{teixeira2019adaloss}. This strategy has a substantial limitation since it only decreases the standard deviation using the MSE loss variances rate without considering the impact of standard deviations changes on MSE loss explicitly. The non-increasing update restriction with a coarse step used in Adaloss might also lead to a sub-optimal solution. The larger the MSE loss variances rate, the bigger the decrease step, might further harm the detection performance.

In the last two years, meta-learning has become the most promising research field in deep learning as it can acquire knowledge versatility for various computer vision tasks, such as image recognition \cite{zoph2018learning, liu2018progressive}. Automating meta-parameters, such as the standard deviation of Gaussian distributions, is highly desired to improve anatomical landmark detection accuracy in medical images. This paper introduces a novel learning-to-learn framework to optimize the objective metrics together with the network parameters for landmark localization. In this new framework, we equip the bottom-up and top-down CNN structures that are the most common backbone for the heatmap regression network. We advance an RL-based approach to refine the objective metric for each landmark. The RL-based approach can optimize the multivariate Gaussian distributions adaptively, improving the accuracy and efficiency of the localization system concurrently. The right of Fig. \ref{intro} shows the difference between the standard deviation obtained with fixed Gaussian distributions and our proposed method. Also, instead of passively terminating the agent inference, we introduce an early-stop strategy to get a balance between exploration and exploitation, terminating the RL agents with the optimal target precision for each landmark separately. The proposed method is trained and validated on two different landmark localization applications, including our in-house prenatal ultrasound (US) dataset and the publicly available dataset of cephalometric X-Ray landmark detection supported by the IEEE International Symposium on Biomedical Imaging (ISBI) \cite{2016A}. Experiment results demonstrate the effectiveness of our proposed method. Our proposed framework is general and has great potentials to improve the efficiency of anatomical landmark detection.

\section{Methods}
\begin{figure*}[!t]
 \centerline{\includegraphics[width=\textwidth]{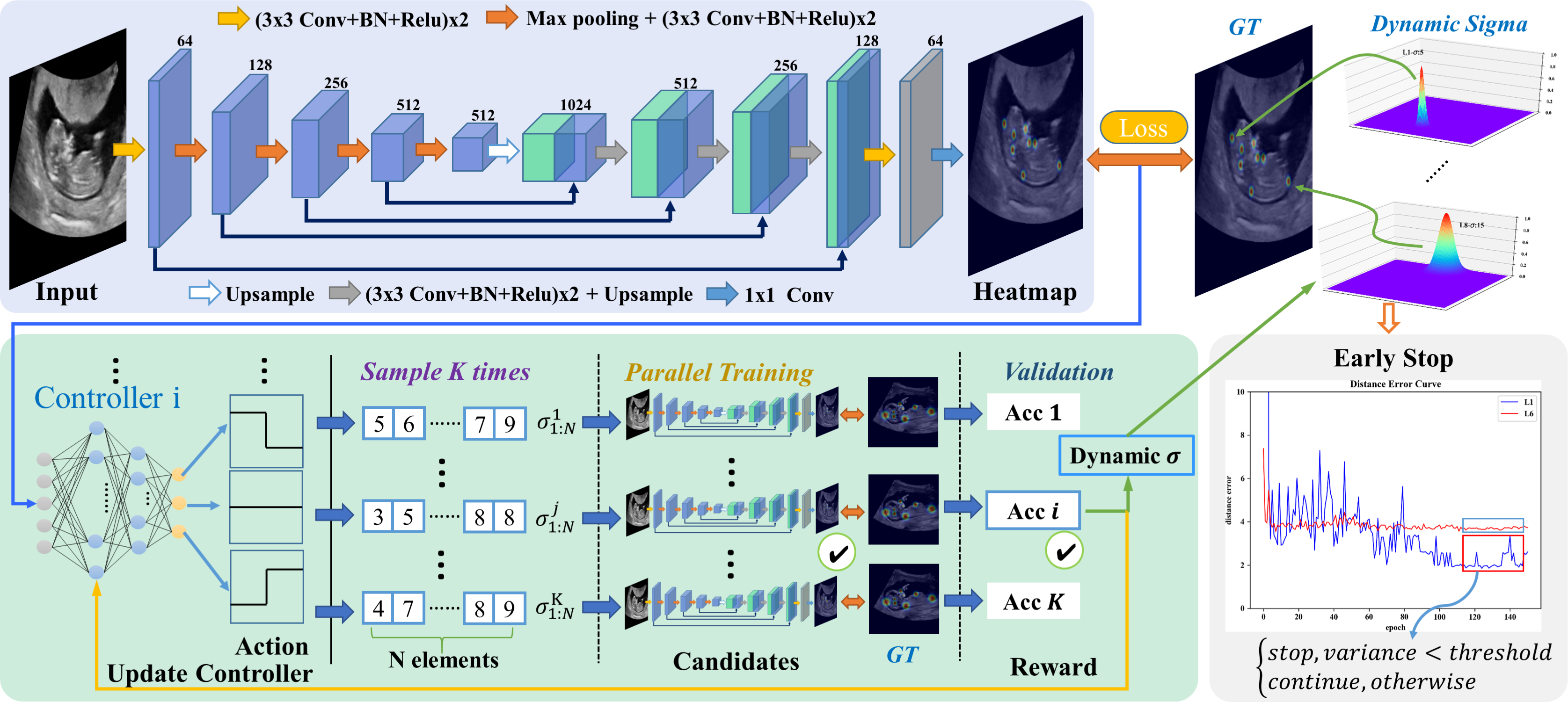}}
 \caption{Schematic view of our proposed learning-to-learn framework. The neural network parameters and the target precision are optimized simultaneously. An RL framework with an early-stop strategy is leveraged to search optimal standard deviation of Gaussian for the generation of ground truth heatmap.}
 \label{framework}
\end{figure*}

Fig. \ref{framework} illustrates the schematic view of our proposed framework. We intend to localize multiple anatomical landmarks in medical images with a learning-to-learn framework. This framework progressively interacts with data information and learns about the optimal network parameters and the corresponding objective metrics. Specifically, we equip the state-of-the-art bottom-up and top-down U-Net \cite{ronneberger2015u} with an RL-based framework \cite{williams1992simple} to adaptively tune the network parameters and Gaussian heatmap variance, thus optimizing the target precision for each landmark. We also design an early-stop strategy for active termination of the interaction procedure to improve its accuracy and efficiency.

\subsection{Landmark-Aware Objective Metric Learning}
The encoder-decoder model is the most common architecture for the heatmap regression network, which regresses heatmaps generated from landmark coordinate using MSE loss. The substance of the network is the multivariate Gaussian distributions centered at the ground truth landmark locations, determining the objective metric of estimating pixel intensity against heatmap. The variance of the Gaussian distributions is, therefore, critical to the accuracy of landmark localization. Although the heatmap regression prefers a small possible standard deviation for better localization accuracy, it generates the target heatmaps (foreground pixels) with highly sparse output distributions. This sparse distribution might be attributed to background pixels (pixels with zero values) that dominate foreground pixels on a heatmap, significantly increasing network training difficulty. Larger variances can alleviate this problem with a blurry and dilated heatmap with low intensity. However, it hurts the ability to locate landmarks accurately. Besides, there is an inherited uncertainty and inconsistency among landmarks in medical images. It is almost impossible to explore specific Gaussian distributions for various landmarks and tasks with hand-crafted heuristics, especially when the number of landmarks increases. Inspired by the work \cite{li2019lfs}, we can consider the variances or standard deviations of the Gaussian distributions as the network hyperparameters to determine the distance metric between the predicted values and the ground truth. We propose a learning-to-learn framework equipping U-Net with an RL-based approach, which can steadily optimize the network parameters and the Gaussian standard deviations of the heatmap ground truth simultaneously.

It is challenging to tune the standard deviations of multivariate Gaussian distributions dynamically due to the agnostic landmark deviation. This intricate task can be well modeled under an RL framework, where an agent, in its current state $\bm{S}$, interacts with the environments $\bm{E}$ by making successive actions $\bm{a}\in\bm{A}$ that maximizes the expectation of reward. We, therefore, advance an RL-based landmark-aware objective metric learning method (Fig. \ref{framework}), denoted as \textbf{\textit{LaOML}}. The details of each component are defined as below:

\textbf{Agent}: Each landmark has a certain agent for heatmap optimization, where each agent samples from the multinomial distribution to achieve the appropriate action by employing the Multi-Layer Perceptron (MLP) consisting of 4 layers of nodes, including 5 inputs, 3 outputs, and two hidden layers containing 64 and 32 hidden units. Since the MLP interacts with the environment to seek the optimal objective metric for tuning the Gaussian distributions of each landmark, we name it the controller. To be specific, each MLP controller takes the MSE loss for the relevant landmark calculated in the latest few epochs as the input and predicts the corresponding action probability for heatmap optimization.

\textbf{Environment and State}: we take medical images as the environment and define states as the multivariate Gaussian distributions.

\textbf{Action}: Action is defined as a cumulative adjustment to the value of the standard deviation. Like Adaloss \cite{teixeira2019adaloss}, we optimize the standard deviation, ${\sigma _i}$, for the target of $i$th landmark with a simple additive equation $\sigma _i^t + \Delta \sigma _i^t$. The action space $\Delta \sigma _i^t$ is then defined as $\left[ { - 1,0, + 1} \right]$, whose probabilities are determined with softmax to the MLP controller outputs, which indicates the decrement, preserving, or increment of the standard deviation. For example, given the output probability $\left\{ {0.7,0.2,0.1} \right\}$, the agent has a likelihood of 0.7 to select the '-1' action. The softmax operation is performed for normalization to the MLP controller outputs.

\textbf{Reward}: The reward signal evaluates the actions to indicate what policy the agent should adopt to select the appropriate action. In this study, the reward is calculated based on the localization accuracy. The smaller the localization error is, the higher the reward is. Each valid action gets its scalar reward to indicate whether the controller is moving towards the preferred target. As the controller interacts with $\bm{E}$ to maximize the rewards using localization accuracy, the system can adaptively adjust the corresponding standard deviation ${\sigma _i}$.

This optimization task is a standard bilevel optimization problem, in which we aim to both maximize the reward w.r.t. the controller parameters $\left\{ \bm{\alpha}  \right\}$, and minimizing the loss of the network w.r.t. network parameters $\left\{ \bm{\omega}  \right\}$. We then define the bilevel optimization problem as below:
\begin{gather}
\mathop {\max }\limits_{\bm{\alpha}}  R\left( {\bm{\alpha}}  \right) = r\left( {{M_{{{\bm{\omega}} ^{\rm{*}}}\left( {\bm{\alpha}}  \right)}};{D_v}} \right) \notag \\
{\rm{s}}.{\rm{t}}.{\rm{\;}}{{\bm{\omega}} ^{\rm{*}}}\left( {\bm{\alpha}}  \right) = \mathop {{\rm{argmin}}}\limits_{\bm{\omega}}  {L^{\bm{\alpha}} }\left( {{M_{\bm{\omega}};{D_t}}} \right)
\end{gather}
where $r\left( {{M_{{{\bm{\omega}} ^{\rm{*}}}\left( {\bm{\alpha}}  \right)}};{D_v}} \right)$ means the reward calculated on the validation dataset ${{D_v}}$ with model ${{M_{{{\bm{\omega}} ^{\rm{*}}}\left( {\bm{\alpha}}  \right)}}}$ and ${L^{\bm{\alpha}} }\left( {{M_{\bm{\omega}}};{D_t}} \right)$ indicates the MSE loss function calculated on the training dataset ${{D_t}}$. Both parameters $\left\{ \bm{\alpha}  \right\}$ and $\left\{ \bm{\omega}  \right\}$ are updated alternately. As illustrated in Fig. \ref{framework}, at the inner level, at the $t$th epoch, we sample $K$ times with the corresponding $\left\{ {\sigma _{1:N}^{t,1:K}} \right\}$ of $N$ landmarks to generate $K$ parallel identical heatmap regression networks, and train these networks independently on the training set for $t'$ epochs to optimize their network weight parameters, respectively. We then identify the model with the best localization accuracy as the initial U-Net network at the $\left( {t + t'} \right)$th epoch. Meanwhile, at the outer level, the controllers exploit the reward of the above $K$ parallel optimized networks on the validation set to update the corresponding controller parameters $\left\{ {{{\bm{\alpha}} _i}} \right\}$. We assume that each controller is independent, and then use the REINFORCE rule \cite{williams1992simple} to iteratively update the $i$th controller with the localization accuracy rewards as follows:
\begin{gather}
{\bm{\alpha}} _i^{t + t'} {\approx} {\bm{\alpha}} _i^t + {\eta _i}\frac{1}{K}\mathop \sum \limits_{j = 1}^K {R\left( {{{\bm{\sigma}} ^{t,j}}} \right)} \cdot {\nabla _{\bm{\alpha}}}{\rm{log}}\left( {g\left( {{{\bm{\sigma}} ^{t,j}}} \right)} \right),{{{\bm{\sigma}} ^{t,j}}\sim g}\notag \\
= {\bm{\alpha}} _i^t + {\eta _i}\frac{1}{K}\mathop \sum \limits_{j = 1}^K {R\left( {{{\bm{\sigma}} ^{t,j}}} \right)} \cdot {\nabla _{\bm{\alpha}} }\mathop \sum \limits_{i = 1}^N {\rm{log}}\left( {p\left( {{\bm{\sigma}} _i^{t,j}} \right)} \right),{{{\bm{\sigma}} ^{t,j}}\sim g}
\end{gather}
where ${\eta _i}$ is the learning rate, and $g\left( {{{\bm{\sigma}} ^{t,j}}} \right) = \mathop \prod \limits_{i = 1}^N p\left( {\sigma _i^{t,j}} \right)$ is the joint probability distribution of $N$ landmarks. {$R\left( {{{\bm{\sigma}} ^{t,j}}} \right)$ is the reward of trajectory ${{\bm{\sigma}} ^{t,j}}$}. In this study, instead of combining the outputs of all the controllers as a joint probability distribution, we heuristically update each controller only with their own reward {$R\left( {\sigma _i^{t,j}} \right)$} as follows:
\begin{gather}
{\bm{\alpha}} _i^{t + t'} = {\bm{\alpha}} _i^t + {\eta _i}\frac{1}{K}\mathop \sum \limits_{j = 1}^K {{R\left( {\sigma _i^{t,j}} \right)}} \cdot {\nabla _{\bm{\alpha}} }{\rm{log}}\left( {{p\left( \Delta \sigma _i^{t,j} \right)}} \right) \notag \\
{{R\left( {\sigma _i^{t,j}} \right)} = C - {\epsilon _{i,j}}}
\end{gather}
where ${\bm{\alpha}} _i^t$ is the weight parameters of the $i$th controller at the $t$th epoch, and {$R\left( {\sigma _i^{t,j}} \right)$} denotes the reward of the $i$th controller in the $j$th sample. $C$ is a constant, and $\epsilon _{i,j}$ is the localization error of the landmark $i$ for the $j$th network on the validation set. {Based on the upper bound of the mean localization error in pixel on the validation dataset under the fixed Gaussian heatmap, we empirically set $C$ as 25 to punish those standard deviations whose localization error was larger than 25 pixels with a negative reward.} We alternately train this bilevel model, finally achieving the optimal objective metric without additional training, and reducing the size of the search space from ${3^N}$ as a total to 3 for each landmark. It is worth noting that we update $\bm{\alpha}$ and $\bm{\sigma}$ at the same frequency, implying the consistency between the inputs to the RL-based controller and the multivariate Gaussian distribution. This mechanism avoids the influence of the $\Delta \sigma _i$ on the MSE loss calculation, thus obviating the undesired bias that is one of the shortcomings of adaptive loss \cite{teixeira2019adaloss}.
\begin{figure*}
 \centering
 \includegraphics[width=0.8\textwidth]{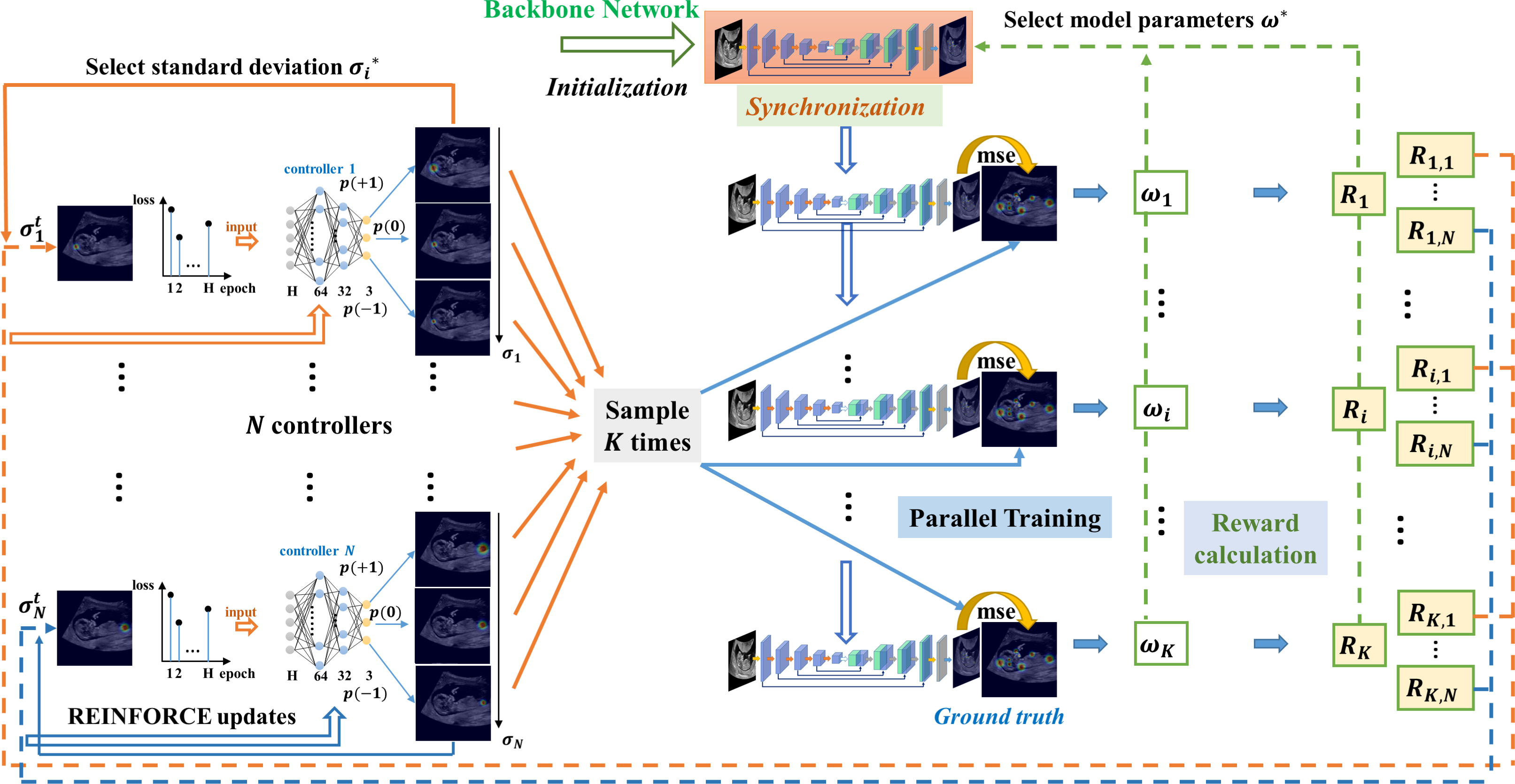}
 \caption{Illustration of RL-based landmark-aware objective metric leaning. {In each optimization iteration ($t'$ epochs), we sample $K$ times according to the output probabilities of each controller to get $K$ sets of standard deviations. $K$ identical U-Net with the same initial parameters are then trained under one standard deviation set in parallel. After that, the best model for initialization in the next iteration is selected and the controllers are updated according to the model performances evaluated on the validation set.}} \label{LaOML}
\end{figure*}
\subsection{Sampling and Broadcasting in \textbf{\textit{LaOML}}}
Fig. \ref{LaOML} shows the proposed \textbf{\textit{LaOML}}. {The sampling and the broadcasting are the two most critical components for parallel models in LaOML to alleviate the exploitation-exploration dilemma that is general in the RL-based framework. In this study, broadcasting the model and actions with the highest validation performance is a kind of exploitation while using $K$ samples can sustain necessary exploration for other actions during training, thus acquiring the well-learned policy. Therefore,} considering action space consists of $\left[ { - 1,0, + 1} \right]$, we employ the multinomial distribution instead of the binomial distribution to model the controller output in the sampling process, indicating the probability of each action to be selected. After one sampling for all the controllers, we get a set of standard deviations $\left\{ {\sigma _{1:N}^{t,j}} \right\}$, which corresponds to the corresponding heatmap of each landmark at the $t$th epoch in the $j$th sample. Thus, we can acquire $\left\{ {\sigma _{1:N}^{t,1},\sigma _{1:N}^{t,2}, \cdots ,\sigma _{1:N}^{t,K}} \right\}$ after $K$ samples for generating $K$ identical U-Net networks, with each U-Net outputting $N$ heatmaps for $N$ landmarks.

Moreover, another essential concern is to train and synchronize identical U-Net networks with different sets of Gaussian distributions. Specifically, we initialize all independent networks with the same weights. Simultaneously, we assign different heatmaps, $\left\{ {\sigma _{1:N}^{t,K}} \right\}$, to train these U-Net networks separately. After training, all parallel models duplicate the model that achieves the best localization performance on the validation set. This broadcast strategy ensures all parallel U-Net networks with the same weights for the next iteration of searching. Moreover, we seek the highest reward among different sample settings and choose the corresponding $\sigma$ as the initial value for each landmark in the next iteration. This sampling and broadcasting procedure is repeated until the end of the whole LaOML training.

\subsection{Early-Stop Strategy}
The continuous change in the multivariate Gaussian distributions is conducive to accurately localizing landmarks. It alleviates the training problem with highly sparse output distributions by dynamically balancing the background pixels and foreground pixels on a heatmap. Nonetheless, an appropriate standard deviation, $\sigma$, is still crucial for the more stable and effective landmark localization. We prefer to stop searching the standard deviations in the proper time, thus fully exploiting the knowledge learned by the RL approach and reducing the search space to optimize. However, there are no well-defined criteria to terminate the iterative inference of RL learning. We propose an early-stop strategy to tell the controller when to immobilize the standard deviation of Gaussian mode during the training, precluding the exploration-exploitation dilemma and realizing an efficient interaction of the RL controller.

An essential issue of an automatic early-stop is to choose an indicator that can accurately reflect the training progress of each landmark. In this study, we select the variance of distance error between the predicted position and the ground truth as a reliable indicator to restrain the change in Gaussian distribution. A small enough variance suggests that the optimization in Gaussian distributions achieves a local extremum and has little effect on the corresponding landmark localization. We, therefore, should stop the search in Gaussian distributions space, vice versa. Specifically, we will record the distance error of each landmark and calculate their variances ${v_{{\rm{DE}},{\rm{\;i}}}}$ on the validation set using the latest $M$ epochs. As illustrated in the bottom right part of Fig. \ref{framework}, the early-strop indicator ${S_i}$ is then determined for the $i$th landmark with a predefined threshold ${{\rm{T}}_{\rm{s}}}$, using:
\begin{equation}
{S_i}=
\begin{cases}
True& {v_{{\rm{DE}},{\rm{\;i}}}} < {{\rm{T}}_{\rm{s}}}\\
False& otherwise
\end{cases}
\end{equation}

If the variance is smaller than ${{\rm{T}}_{\rm{s}}}$, the indicator is set to true for immobilizing $\sigma _i$ for the Gaussian centered at the $i$th landmark coordinate. Otherwise, the RL-based controller continues to tune the corresponding standard deviation. In this study, the $M$ is set to be 30 epochs. Also, we empirically determine the threshold ${{\rm{T}}_{\rm{s}}}$ to be 0.01 by calculating the variance from distance errors within 30 epochs achieving the stable landmark localization in our pre-experiment. Moreover, to sufficiently explore a higher probability of finding the proper $\sigma$, we employ the early-stop strategy after the first 100 epochs.

\section{Experiments}
\subsection{Datasets}
We assessed the localization performance of the proposed method on two different datasets. The first one is an in-house prenatal ultrasound dataset of 511 fetal midsagittal plane image with a size of 400$ \times $400. All images were anonymized and obtained by experts from 3D fetal ultrasound volumes acquired from three ultrasound scanners (GE, Philips, and Mindray). A clinical expert provided ground truth of 8 landmarks for all the images, including crown, diencephalon, thalamus, nasal bone, inferior alveolar bone, back neck point, anterior third of the chest wall, and rump. The local Institutional Review Board approved this study. The second dataset is a publicly available dataset from the Automatic Cephalometric X-Ray Landmark Detection Challenge (ACXRLDC) supported by the ISBI in 2015 \cite{2016A}. The ACXRLDC dataset consists of 400 lateral cephalograms with 1935$ \times $2400 and a resolution of 0.1 mm/pixel, which is split randomly into three subsets: 150 Training images, 150 Test1 images, and 100 Test2 images. Two experts manually marked 19 cephalometric landmarks for all images in the ACXRLDC dataset. The mean value of the manual annotations is used as the ground truth of each landmark.

 We split the in-house prenatal ultrasound dataset into 60$\%$ training, 20$\%$ validation, and 20$\%$ test, randomly. By contrast, we follow dataset usage in \cite{chen2019cephalometric} to use Training images, Test1 images, and Test2 images for training, validation, and test, respectively. Different augmentation strategies were applied for both datasets, including flipping, shifting scaling, rotation, and changing brightness and contrast. All images were resized to 256$ \times $256 for the prenatal ultrasound dataset and 800$ \times $640 for the ACXRLDC dataset, respectively. During testing, the locations of landmarks were rescaled back to the original resolution for the metric evaluation.

\subsection{Implementation details}
We evaluated our proposed method based on the U-Net backbone, denoted as the baseline model (Fig. \ref{framework}). We compared our proposed LaOML with two other different strategies of refining the predicted output, generative adversarial learning (GAN) \cite{xu2018less,chen2017adversarial}, and recurrent neural network (RNN) \cite{liu2019feature}. Moreover, we also explore the method \cite{jakab2018unsupervised} to extract the exact pixel's coordinate directly from the predicted heatmap.

We reimplemented all these methods using the same U-Net backbone and MSE loss in our codes with Pytorch to make them compatible with our experiments. A ResNet34 was utilized as a discriminator to distinguish the predicted heatmaps and the ground truth in the GAN method, where the heatmaps and the original image were taken as the input of the discriminator branch. By contrast, in the RNN method, a convolutional RNN layer was added on top of the backbone to explore the dependencies among different landmarks' heatmaps. According to \cite{jakab2018unsupervised}, the predicted heatmap can be renormalised to a probability distribution via (spatial) Softmax and condensed to exact coordinate as below:
\begin{equation}
{{u_i} = \frac{{\mathop \sum \nolimits_{u \in {\rm{\Omega }}} u \cdot {e^{{S_u}\left( i \right)}}}}{{\mathop \sum \nolimits_{u \in {\rm{\Omega }}} {e^{{S_u}\left( i \right)}}}}}
\end{equation}
where ${\rm{\Omega }}$ represents the image domain. ${u_i}$ denotes the predicted coordinate, and ${S_u}\left( i \right)$ means the heatmap output for the $i$th landmark. The MSE loss between the predicted coordinates and ground truth is then used to train the backbone. We denote our baseline model as the BASE, while denoting the baseline model equipped with LaOML, GAN, and RNN as the BASE-LaOML, the BASE-GAN, and the BASE-RNN, respectively. The method based on direct coordinate regression is denoted as BASE-C. Moreover, we also evaluated the proposed method with the results in state-of-the-art literature about the public ACXRLDC dataset \cite{chen2019cephalometric, arik2017fully, song2020automatic, ibragimov2015computerized, lindner2015fully}.

We ran all experiments on a machine with an NVIDIA TITAN X(PASCAL) GPU. We ensured to set the initial parameters appropriately. We trained U-Net network parameters using Adam optimizer with learning rate 2e-4 while updating the controller weights of LaOML with Adam optimizers and learning rates 1e-3. We set the sample number $K=10$ and update the controllers with $C=25$ every 5 epochs when employing LaOML. The vector length of the controller input was set to be 5 as well correspondingly, which is equal to the update interval of the controller. We searched $\sigma$ ranged from 1 to 20 and stopped the search with the early-strop indicator ${S_i}<0.01$. After 40 epochs for the in-house ultrasound dataset and 30 epochs for the ACXRLDC dataset, we began to search parameters initialized with 5 for all landmarks, improving the training stability. Moreover, we achieved the best performance of the BASE-C method with a learning rate of 1e-3 and 1e-2 on the prenatal ultrasound dataset and ACXRLDC dataset. By contrast, we updated the RNN model using Adam optimizer with a learning rate of 2e-4 and 2e-5 on the prenatal ultrasound dataset and ACXRLDC dataset, respectively. On the other hand, we trained the GAN model using Adam optimizers with learning rates of 1e-4 and 4e-4 to update the heatmap regression model and discriminator, respectively, for the prenatal ultrasound dataset, while the corresponding learning rates for the ACXRLDC dataset are 2e-5 and 8e-4 to get better performance. We trained all models for 250 epochs. Following the optimization configuration of previous meta-learning methods \cite{li2019lfs, zoph2016neural, baker2016designing}, we employ the training dataset and validation dataset to optimize the network parameters and the corresponding network hyperparameters (i.e. the Gaussian distributions' standard deviations), respectively. During inference, the pixel with the highest value on each channel of the output heatmap was selected as the final predicted position of each landmark through an argmax function.

\subsection{Evaluation criteria}
We adopt two frequently used metrics to quantitatively evaluate landmark localization accuracy, mean radial error (MRE), and the standard Percentage of Correct Keypoints (PCK) metric.
\begin{equation}
{\rm{MRE}}\left( j \right) = \frac{{\sum \Delta {X_{i,j}}}}{{\# {D_{test}}}}
\end{equation}
where $\Delta {X_{i,j}}$ is the Euclidean distance (mm) between the predicted position and the ground truth for the $j$th landmark in the $i$th testing image, and ${\# {D_{test}}}$ is the total number of the testing images. We also calculate the Mean ($ \pm $S.D.) MRE of all landmarks when evaluating our methods.

The PCK, also, namely success detection rate (SDR), reports the percentage of detections with Euclidean distance below a threshold. It identifies how many target landmarks can be successfully localized considering a certain margin of error
\begin{equation}
{\rm{PC}}{{\rm{K}}_r} = \frac{{\# \left\{ {\left( {i,j} \right):\Delta {X_{i,j}} < r} \right\}}}{{N \times \# {D_{test}}}} \times 100{\rm{\% }}
\end{equation}
with ${N \times \# {D_{test}}}$ is the number of all the localized landmarks, and $\# \left\{  \cdot  \right\}$ denotes the amount of the localized landmarks which satisfy the condition $\left\{  \cdot  \right\}$. We follow the suggestion in the previous literature about the public ACXRLDC dataset \cite{chen2019cephalometric, arik2017fully, song2020automatic, ibragimov2015computerized, lindner2015fully} to use 2mm, 2.5mm, 3mm, and 4mm as the margin of errors when calculating PCK for the ACXRLDC dataset.

\section{Results}
\subsection{Evaluation on the prenatal ultrasound dataset}
\begin{figure}[!htb]
 \centerline{\includegraphics[width=0.8\columnwidth]{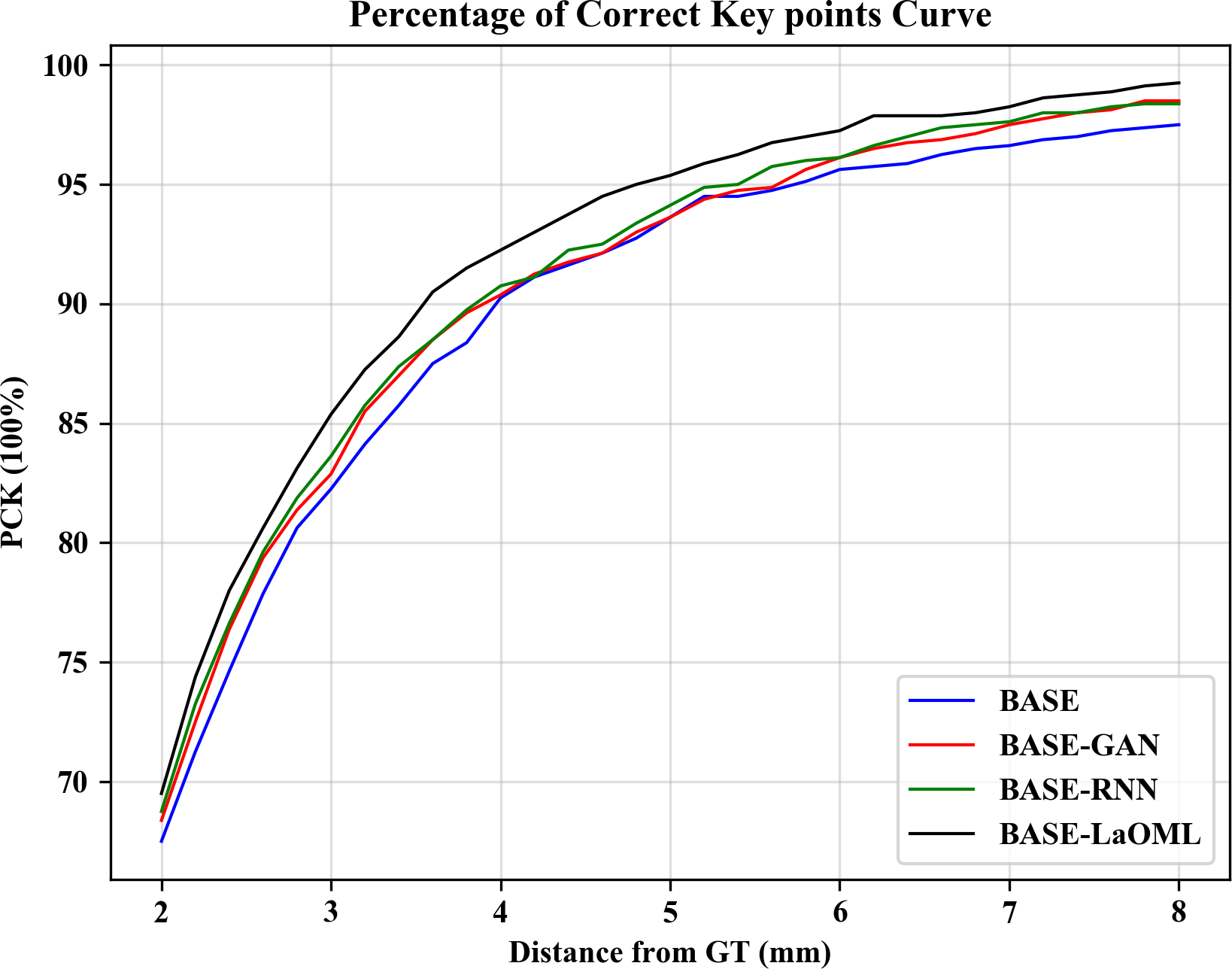}}
 \caption{PCK curve on the prenatal ultrasound dataset.}
 \label{PCK_US}
\end{figure}

\begin{figure}[!htb]
 \centerline{\includegraphics[width=\columnwidth]{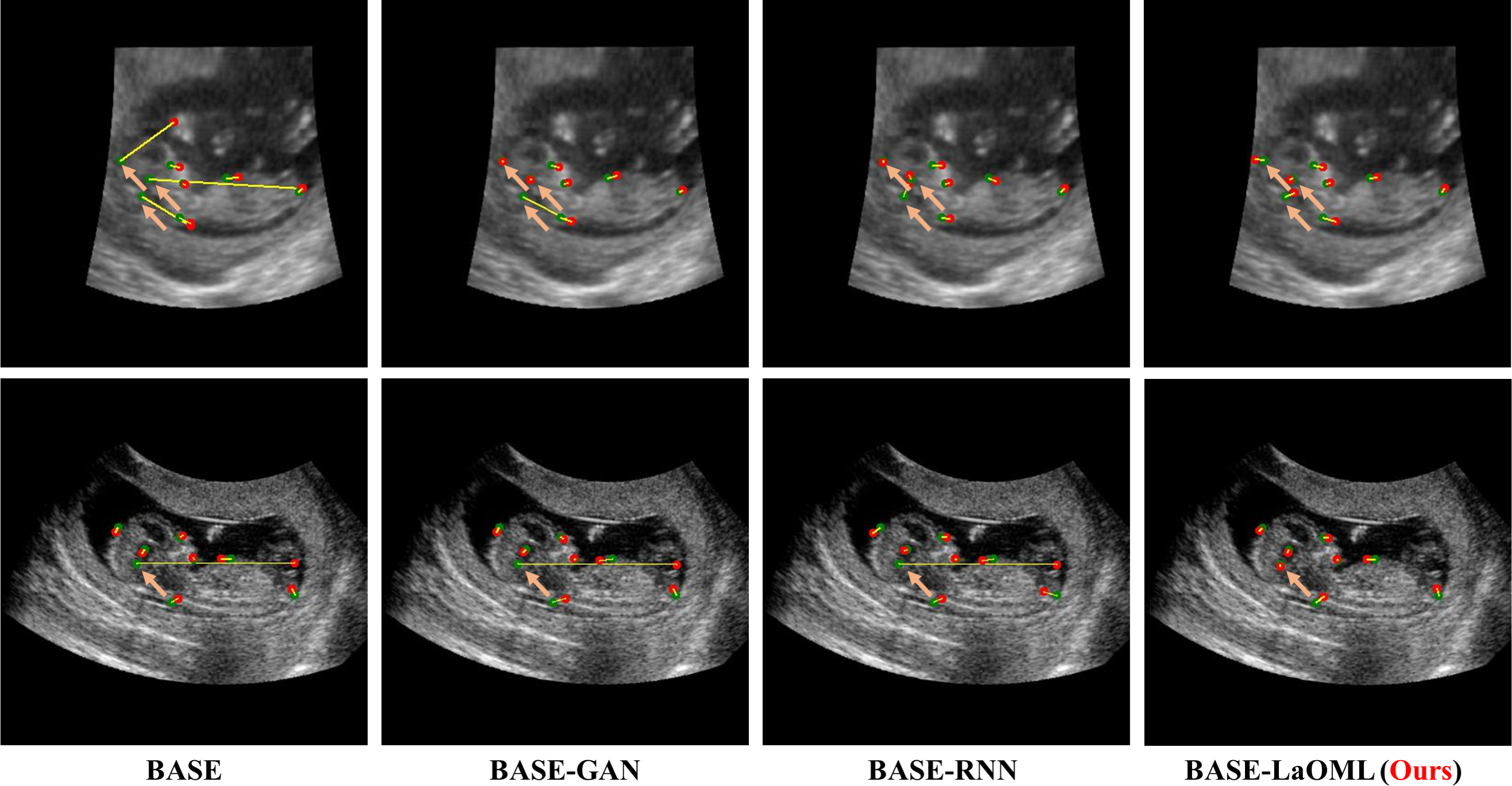}}
 \caption{Examples of extreme situations on the prenatal ultrasound dataset. The images in each column represent different prediction results using a specific method. Each row shows the predictions using different methods on the same image, including BASE, BASE-GAN, BASE-RNN, and BSAE-LaOML. The green and red points represent ground truth and predicted position, respectively.}
 \label{extreme_US}
\end{figure}

\begin{figure*}[!t]
 \centerline{\includegraphics[width=0.8\textwidth]{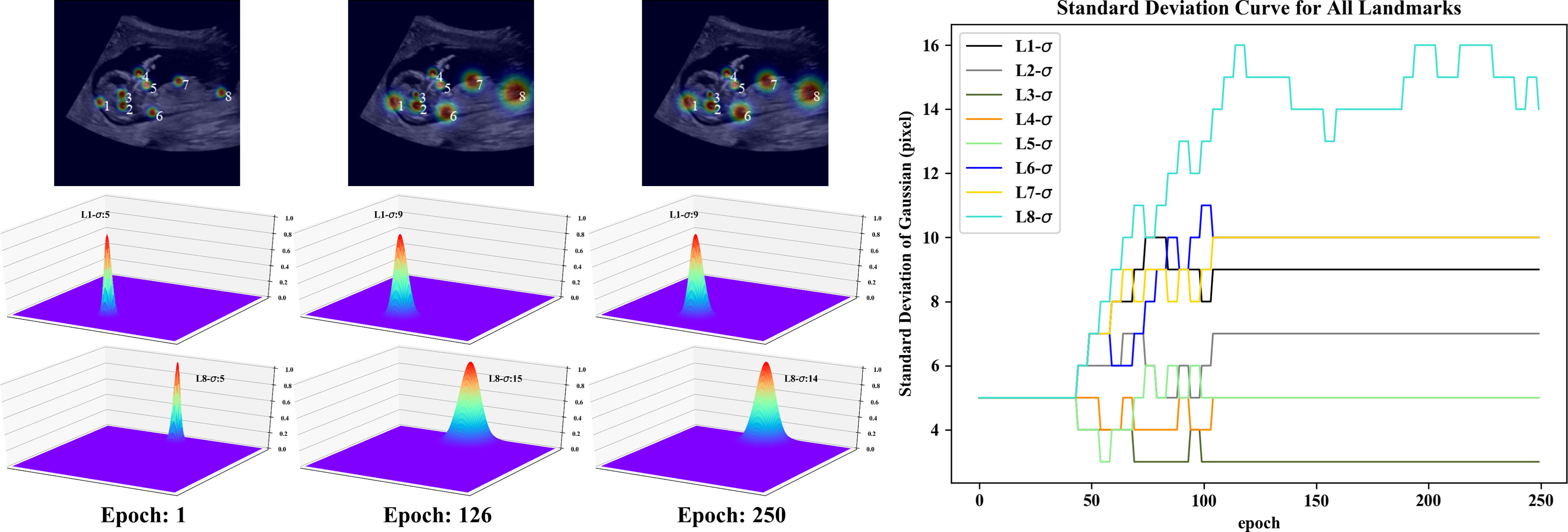}}
 \caption{The dynamic change of Gaussian $\sigma$ on the prenatal ultrasound dataset. The text 'Ln-$\sigma$' indicates the $\sigma$ change curve of the $n$th landmark. In the left three columns, we show the overall Gaussian distributions of all the landmarks on a single image (Row 1) and the specific distribution of the first (Row 2) and the eighth landmark (Row 3) in a 3D visualization along with training epoch. In the right part, the dynamic standard deviation curve for each landmark is presented.}
 \label{sigma_US}
\end{figure*}

Table~\ref{Total Comparison of US}  presents the landmark detection results when different strategies are employed on the BASE network. Our proposed BASE-LaOML significantly outperforms BASE, BASE-GAN, BASE-RNN, and BASE-C, achieving a good improvement in mean MRE of all landmarks. The LaOML attains the lowest MRE average and STD and achieves the greatest improvements of MRE on 6 of 8 landmarks, while GAN and RNN only enhance the baseline to a certain extent. Moreover, the direct usage of the exact pixel's coordinate deteriorates the localization accuracy compared to the heatmap regression methods. Fig. \ref{PCK_US} shows the curve between PCK and the threshold ranging from 2 to 8 mm for the heatmap regression methods. The BASE-LaOML also tops the task and achieves the highest PCK result, outperforming BASE by 2$\%$ under almost all the thresholds. Both BASE-GAN and BASE-RNN also show some improvements in PCK. Nevertheless, the BASE-LaOML still outperforms these two methods in both MRE and PCK (Table \ref{Total Comparison of US} and Fig. \ref{PCK_US}). We also visualize some examples of the localized landmarks for the prenatal ultrasound (Fig. \ref{extreme_US}). The results suggest that using LaOML restrains the extreme errors of landmarks' localization. However, GAN or RNN cannot suppress these errors effectively, though they can alleviate this problem. Moreover, Fig. \ref{sigma_US} presents the process of optimizing $\sigma$ using our proposed LaOML method, implying the inherited uncertainty and inconsistency in landmark heatmap. These qualitative and quantitative results suggest that our proposed method is comparatively more robust to speckle noise and achieves proper localization in ultrasound images.

\begin{table}[!htb] \caption {Comparison of MRE (mm) on the prenatal ultrasound dataset.} \label{Total Comparison of US}
  \centering
  \tiny
  \begin{tabular}{c| c c c c c c c c c}
  \hline
  \bf{Methods} &L1 &L2 &L3 &L4 &L5 &L6 &L7 &L8 &Mean$ \pm $S.D.\\
  \hline
   BASE     &3.04 &2.28 &1.73 &1.52 &1.22 &2.65 &2.99 &2.31 &2.22$ \pm $3.58 \\
   BASE-GAN   &2.87 &2.02 &1.43 &\bf{1.38} &1.02 &3.26 &2.77 &2.71 &2.18$ \pm $4.39 \\
   BASE-RNN   &3.13 &2.03 &\bf{1.20} &1.40 &0.99 &2.77 &\bf{2.60} &1.86 &2.00$ \pm $2.67  \\
   BASE-C   &3.03 &2.37 &1.92 &2.14 &1.80 &3.10 &3.04 &2.92 &2.54$ \pm $1.98  \\
   BASE-LaOML     &\bf{2.65} &\bf{1.35} &\bf{1.20} &1.39 &\bf{0.98} &\bf{2.36} &2.66 &\bf{1.79} &\bf{1.80}$ \pm $\bf{1.55} \\
   \hline
  \end{tabular}
\end{table}

\begin{table}[!htb] \caption {Comparison of MRE (mm) and PCK on the ACXRLDC dataset.} \label{Total Comparison of X}
  \centering
  \tiny
  \begin{tabular}{c| c c c c c}
  \hline
  \bf{Methods} &Mean$ \pm $S.D. $\downarrow$ &$PCK_2$$\uparrow$ &$PCK_{2.5}$$\uparrow$ &$PCK_3$$\uparrow$ &$PCK_4$$\uparrow$  \\
  \hline
   Ibragimov\cite{ibragimov2015computerized} &-    &62.74 &70.47 &76.53 &85.11 \\
   Lindner\cite{lindner2015fully}   &-    &66.11 &72.00 &77.63 &87.42 \\
   Arik\cite{arik2017fully}      &-    &67.68 &74.16 &79.11 &86.63 \\
   Chen\cite{chen2019cephalometric}      &1.48 &75.05 &82.84 &88.53 &\bf{95.05} \\
   Song\cite{song2020automatic}      &1.54 &74.0  &81.3  &87.5  &94.3  \\
   BASE      &1.57$ \pm $3.30 &73.16 &81.26 &86.42 &93.89 \\
   BASE-GAN    &1.49$ \pm $\bf{1.31} &74.16 &81.89 &87.58 &94.26 \\
   BASE-RNN    &1.47$ \pm $1.34 &74.16 &81.63 &87.26 &94.42 \\
   BASE-C    &2.68$ \pm $2.26 &47.11 &57.74 &66.95 &79.11 \\
   BASE-LaOML      &\textbf{1.39}$ \pm $1.32 &\bf{76.11} &\bf{84.21} &\bf{88.79} &94.84 \\
   \hline
  \end{tabular}
\end{table}

\subsection{Evaluation on the ACXRLDC dataset}
\begin{figure*}[!t]
 \centerline{\includegraphics[width=0.7\textwidth]{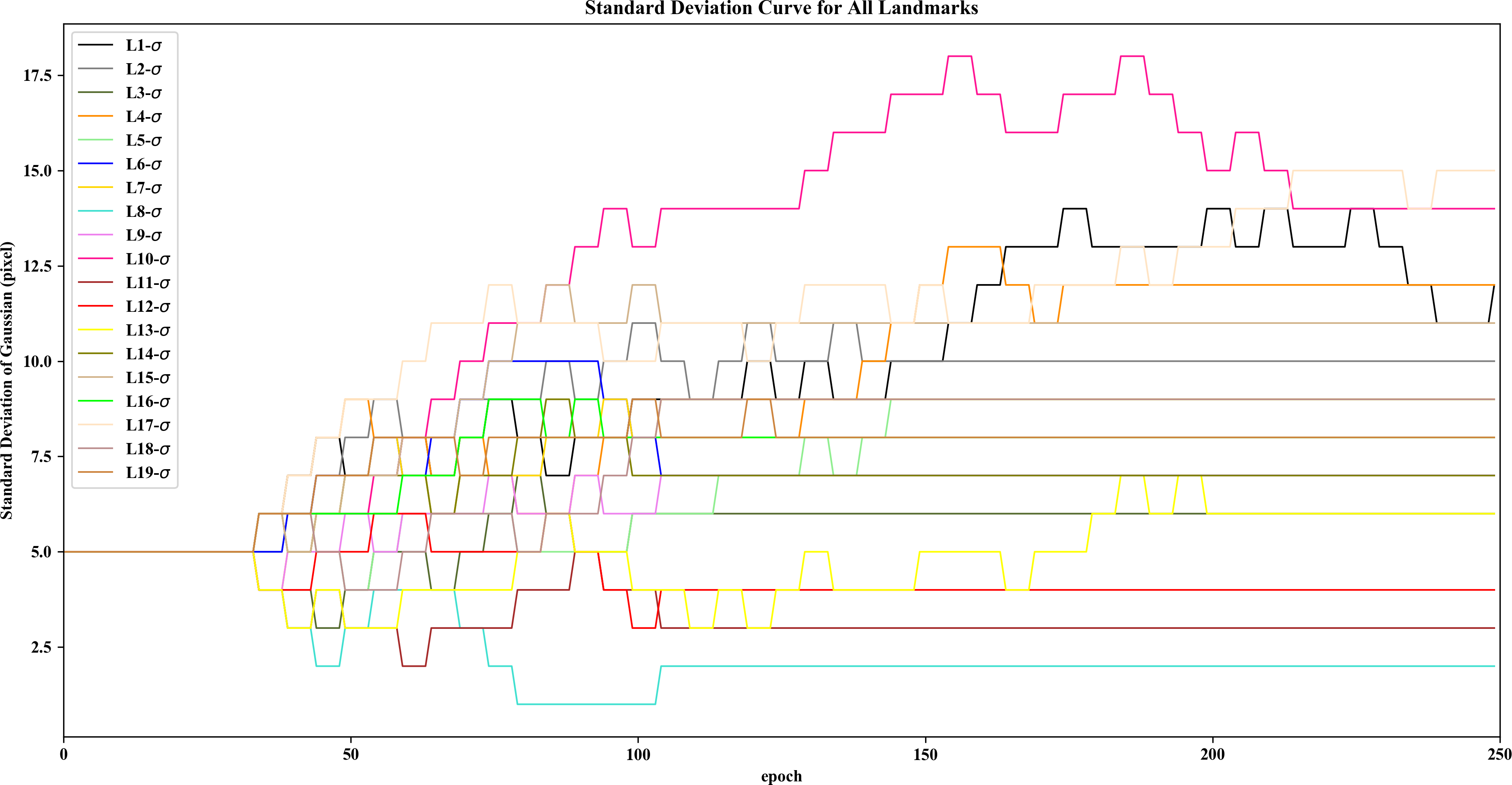}}
 \caption{The dynamic change of Gaussian $\sigma$ on the ACXRLDC dataset. The text 'Ln-$\sigma$' indicates the $\sigma$ change curve of the $n$th landmark. The dynamic standard deviation curve for each landmark is presented.}
 \label{sigma_X}
\end{figure*}

\begin{figure}[!htb]
 \centerline{\includegraphics[width=0.75\columnwidth]{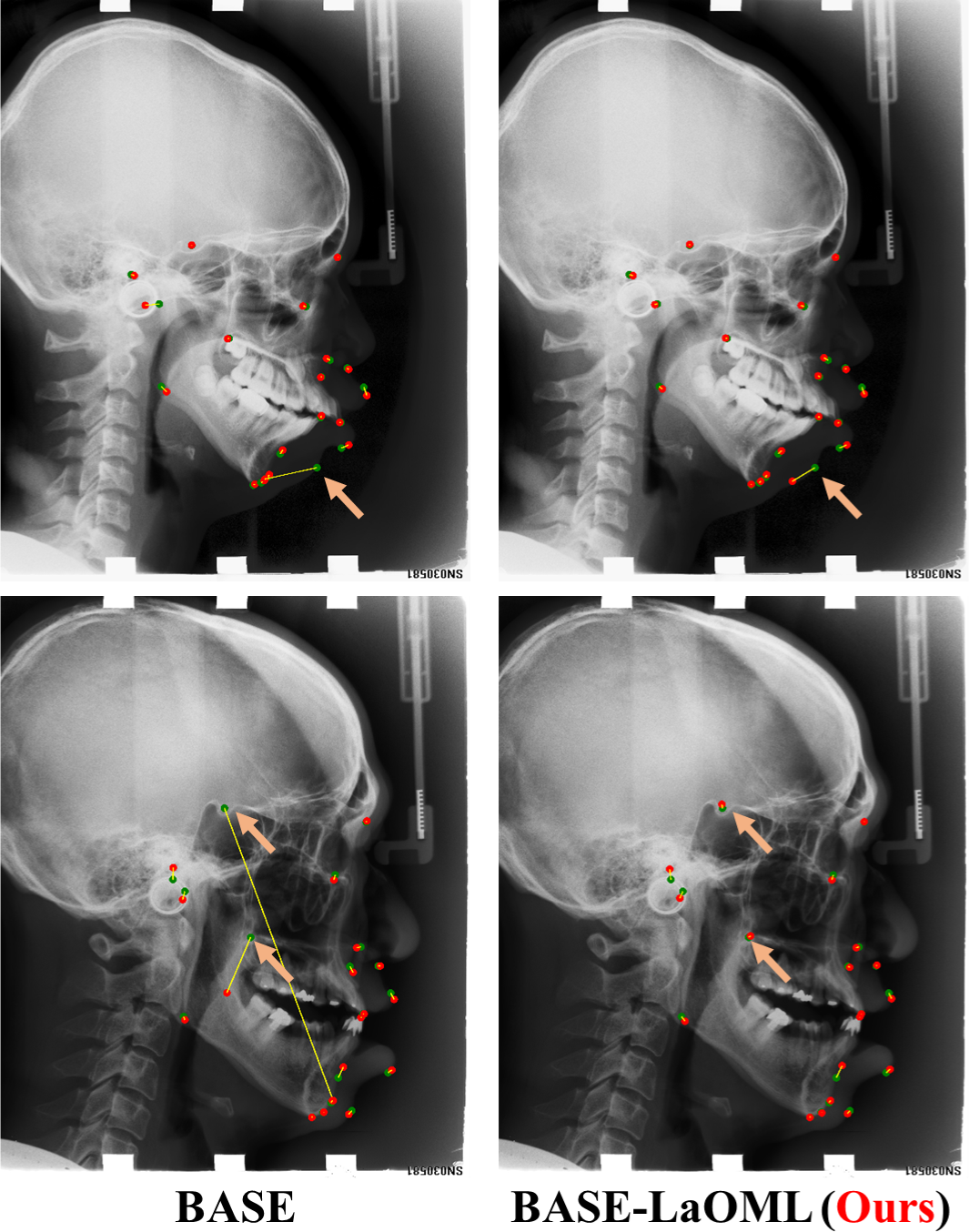}}
 \caption{Examples of extreme situations on the ACXRLDC dataset. The images in each column represent different prediction results using a specific method. Each row shows the predictions using different methods on the same image, including BASE, and BSAE-LaOML. The green and red points represent ground truth and predicted position, respectively.}
 \label{extreme_X}
\end{figure}

We show the results of comparing our BASE-LaOML to BASE-RNN, BASE-GAN, BASE-C, and other state-of-the-art methods on the Test2 subset of the ACXRLDC dataset in Table \ref{Total Comparison of X}. We offer the PCK under 2 mm, 2.5 mm, 3 mm, and 4 mm threshold and the Mean ($ \pm $S.D.) MRE. Our baseline model obtains only a little worse result than most state-of-the-art methods, although it does not make additional network constraints. As illustrated in Fig. \ref{sigma_X}, landmarks in Cephalometric X-Ray images prefer the different $\sigma$ to generate the heatmap ground truth. We conjecture that choosing an appropriate $\sigma$ instead of a commonly small $\sigma$ can improve the model. We validate this assumption in the following ablation studies by using different $\sigma$. Compared to the BASE, the BASE-LaOML method performs preferably in landmark localization with MRE improvement of 0.18 mm in average and 2 mm in standard deviation. When comparing with other state-of-the-art algorithms, it is clear that the average MRE for most of the other results is above 1.45mm. Also, in line with the prenatal ultrasound dataset results, the Base-C method has the worst localization performance among all the compared methods. Fig. \ref{extreme_X} presents some examples from different patients of the ACXRLDC dataset. Most predicted landmarks correspond to the positions where they are supposed to be. Also, the results support that the proposed LaOML alleviates the extreme errors in landmarks' localization.

In terms of PCK to evaluates the distribution of predicted landmarks around ground truth, the BASE-LaOML achieves the best performance on other metrics, although our methods are only slightly less than the results in \cite{chen2019cephalometric} when the threshold of PCK is set to 4.0 mm. Additionally, Payer et al. \cite{payer2019integrating} present results under the other arrangement by combining two test datasets. Nevertheless, the attentive feature pyramid fusion (AFPF) introduced in \cite{chen2019cephalometric} has achieved much higher accuracy under the same configuration. These results suggested that our BASE-LaOML is more accurate in localization and the same robust to mislocalization as the AFPF \cite{chen2019cephalometric}. Moreover, all localization improvements can be attributed to additional geometric constraints for the anatomically ill-defined landmarks, such as attention mechanism \cite{chen2019cephalometric}, local patches \cite{song2020automatic}, and learnable $\sigma$. However, both GAN and RNN methods gain fewer improvements in comparison with other constraints methods. This can be partly explained as GAN makes the heatmap output more consistent with the prior heatmap, but it poses no localization constraints directly to reduce localization error. On the other hand, the RNN-based model might prefer a more complicated and task-specific relation among landmarks to fully explore the dependence among different landmarks, restricting the positioning accuracy.

\subsection{Ablation experiments}
We run a number of ablation studies to analyze the localization performance of different fixed $\sigma $ and the effect of initial values of the learnable $\sigma$ on the learning process and model performance. Besides, we investigate the efficiency of our proposed early-stop strategy. The results are discussed in detail next.
\subsubsection{Fixed standard deviation parameters $\sigma $} The ablation study in Table \ref{Detail Comparison of US} and Table \ref{Detail Comparison of X} shows the BASE model results with different fixed $\sigma$ from 3 to 20. It is clear that varying $\sigma $ results in the difference in localization accuracy, reaching more than 0.5 mm discrepancy for some landmarks such as L1, L4, L16, L17, and L19 (Table \ref{Detail Comparison of X}). These landmarks are usually anatomically ill-defined (e.g., L16, the anterior landmark at the chin in Fig. \ref{extreme_X}). As shown in Table \ref{Detail Comparison of US} and Table \ref{Detail Comparison of X}, it is uncorrelated between the localization accuracy and the value of $\sigma$. Each landmark prefers to independent heatmap with specified peak widths $\sigma $, suggesting the need for learnable $\sigma $ to adapt to the data by encoding network prediction uncertainty. It is interesting to notice that larger fixed $\sigma$ outperforms the results with smaller $\sigma$. As demonstrated in Fig. \ref{sigma_US} and Fig. \ref{sigma_X}, most landmarks prefer $\sigma$ larger than 5, though one or two landmarks learn the $\sigma$ to 3. Thus, we develop the BASE model initialized with $\sigma {\rm{ = }}5$ in other experiments, making a comprehensive and fair comparison of the learnable $\sigma$ using LaOML.

\begin{table}[!htb] \caption {Comparison of MRE (mm) for the BASE model with different fixed $\sigma$ on the prenatal ultrasound dataset.} \label{Detail Comparison of US}
  \centering
  \tiny
  \begin{tabular}{c| c c c c c c c c c}
  \hline
  \bf{Methods} &L1 &L2 &L3 &L4 &L5 &L6 &L7 &L8 &Mean$ \pm $S.D. \\
  \hline
   BASE-$\sigma$3     &3.77  &2.75 &1.22 &1.56 &1.06 &3.50  &3.56 &3.69 &2.64$ \pm $5.32 \\
   BASE-$\sigma$5     &3.04  &2.28 &1.73 &1.52 &1.22 &2.65  &2.99 &2.31 &2.22$ \pm $3.58 \\
   BASE-$\sigma$8     &2.69  &1.47 &1.25 &1.38 &1.01 &2.36  &2.61 &2.35 &1.89$ \pm $2.30 \\
   BASE-$\sigma$10    &2.56  &1.50 &1.23 &\bf{1.34} &1.08 &2.39  &2.60 &2.09 &1.85$ \pm $1.55 \\
   BASE-$\sigma$15    &2.77  &1.57 &1.21 &1.41 &1.06 &2.47  &\bf{2.44} &2.01 &1.87$ \pm $1.45 \\
   BASE-$\sigma$18    &2.69  &1.57 &1.35 &1.38 &1.19 &2.36  &2.47 &1.99 &1.88$ \pm $\bf{1.37} \\
   BASE-$\sigma$20    &\bf{2.64}  &1.66 &1.31 &1.39 &1.26 &\bf{2.29}  &2.48 &2.17 &1.90$ \pm $1.46 \\
   BASE-LaOML   &2.65  &\bf{1.35} &\bf{1.20} &1.39 &\bf{0.98} &2.36  &2.66 &\bf{1.79} &\textbf{1.80}$ \pm $1.55 \\
   \hline
  \end{tabular}
\end{table}

\begin{table*}[!htb] \caption {Comparison of MRE (mm) for the BASE model with different fixed $\sigma$ on the ACXRLDC dataset.} \label{Detail Comparison of X}
  \centering
  \tiny
  \begin{tabular}{c| c c c c c c c c c c c c c c c c c c c c}
  \hline
  \bf{Methods} &L1 &L2 &L3 &L4 &L5 &L6 &L7 &L8 &L9 &L10 &L11 &L12 &L13 &L14 &L15 &L16 &L17 &L18 &L19 & Mean$ \pm $S.D.\\
  \hline
   BASE-$\sigma$3     &1.43 &0.87 &2.30 &4.81 &1.15 &2.64 &0.67 &0.66 &0.55 &1.77 &\bf{0.74} &0.46 &2.62 &2.03 &0.88 &6.06 &1.60 &1.03 &3.67 &1.89$ \pm $7.28\\
   BASE-$\sigma$5     &1.84 &0.94 &2.18 &1.68 &1.15 &2.90 &0.61 &\bf{0.62} &0.52 &1.87 &0.87 &0.47 &2.75 &1.96 &1.04 &4.72 &1.23 &1.19 &1.23 &1.57$ \pm $3.30 \\
   BASE-$\sigma$8     &\bf{0.53} &0.90 &2.22 &1.59 &1.09 &2.72 &0.64 &0.63 &\bf{0.47} &1.82 &0.83 &0.50 &2.61 &1.93 &0.89 &\bf{4.42} &\bf{1.00} &1.09 &\bf{1.07} &1.42$ \pm $1.41\\
   BASE-$\sigma$10    &0.55 &0.94 &\bf{2.14} &\bf{1.56} &1.13 &2.65 &\bf{0.59} &0.65 &0.49 &1.64 &0.85 &0.53 &2.60 &\bf{1.90} &0.90 &4.66 &1.04 &\bf{1.03} &1.17 &1.42$ \pm $1.35\\
   BASE-$\sigma$15    &0.62 &0.96 &2.23 &1.66 &1.16 &\bf{2.60} &0.63 &0.64 &0.51 &\bf{1.41} &0.95 &0.62 &2.67 &1.94 &0.89 &4.52 &1.01 &1.18 &1.22 &1.44$ \pm $1.29 \\
   BASE-$\sigma$18    &0.57 &0.90 &2.22 &1.73 &1.17 &2.69 &0.69 &0.68 &0.50 &1.55 &0.90 &0.63 &2.57 &1.95 &0.87 &4.51 &1.01 &1.13 &1.19 &1.45$ \pm $1.29 \\
   BASE-$\sigma$20    &0.56 &0.96 &2.22 &1.74 &1.08 &2.81 &0.65 &0.65 &0.54 &1.47 &0.95 &0.69 &2.56 &2.02 &0.87 &4.48 &1.04 &1.24 &1.18 &1.46$ \pm $\bf{1.27} \\
   BASE-LaOML   &0.55 &\bf{0.86} &2.17 &1.63 &\bf{1.07} &2.67 &0.64 &0.63 &0.50 &1.59 &0.78 &\bf{0.44} &\bf{2.48} &1.93 &\bf{0.84} &4.49 &\bf{1.00} &1.11 &1.11 &\textbf{1.39}$ \pm $1.32\\
   \hline
  \end{tabular}
\end{table*}

\subsubsection{RL-based LaOML with different initial $\sigma $} The results show that our RL-based LaOML helps our model improve the detection accuracy. Our LaOML method achieves the smallest mean MRE. As shown in Table \ref{Detail Comparison of X}, learning $\sigma$ independently get the most accurate localization for six landmarks, and only worse by over 0.1 mm than the best result of fixed $\sigma$ for one landmark, L10. Similar observations are presented in Table \ref{Detail Comparison of US} of the prenatal ultrasound data set. On the other hand, the ablation study shows that the discrepancy in localization is small (Table \ref{Initial US} and Table \ref{Initial X}), though the initial value of $\sigma$ affects our RL-based approach. Also, all learnable $\sigma$ methods still surpass other state-of-the-art methods. These results prove that it is essential to predict the heatmap with learnable $\sigma$, making the model easy to train and robust to misidentification.

\begin{table}[!htb] \caption {Comparison of MRE (mm) for BASE-LaOML with different initial $\sigma$ on the prenatal ultrasound dataset.} \label{Initial US}
  \centering
  \tiny
  \begin{tabular}{c| c c c c c c c c c}
  \hline
  \bf{Methods} &L1 &L2 &L3 &L4 &L5 &L6 &L7 &L8 &Mean$ \pm $S.D. \\
  \hline
   BASE-LaOML-$\sigma$5     &2.65 &1.35 &1.20 &1.39 &0.98 &2.36 &2.66 &1.79 &1.80$ \pm $1.55 \\
   BASE-LaOML-$\sigma$8     &2.65 &1.43 &1.16 &1.33 &1.07 &2.46 &2.52 &1.76 &1.80$ \pm $1.42 \\
   BASE-LaOML-$\sigma$10    &2.68 &1.43 &1.19 &1.38 &1.00 &2.44 &2.49 &1.87 &1.81$ \pm $1.39 \\
   BASE-LaOML-$\sigma$15    &2.72 &1.55 &1.23 &1.36 &1.05 &2.39 &2.49 &1.92 &1.84$ \pm $1.43 \\
   \hline
  \end{tabular}
\end{table}

\begin{table*}[!htb] \caption {Comparison of MRE (mm) for BASE-LaOML with different initial $\sigma$ on the ACXRLDC dataset.} \label{Initial X}
  \centering
  \tiny
  \begin{tabular}{c| c c c c c c c c c c c c c c c c c c c c}
  \hline
  \bf{Methods} &L1 &L2 &L3 &L4 &L5 &L6 &L7 &L8 &L9 &L10 &L11 &L12 &L13 &L14 &L15 &L16 &L17 &L18 &L19 &Mean$ \pm $S.D. \\
  \hline
   BASE-LaOML-$\sigma$5     &0.55 &0.86 &2.17 &1.63 &1.07 &2.67 &0.64 &0.63 &0.50 &1.59 &0.78 &0.44 &2.48 &1.93 &0.84 &4.49 &1.00 &1.11 &1.11 &1.39$ \pm $1.32 \\
   BASE-LaOML-$\sigma$8     &0.58 &0.93 &2.23 &1.61 &1.20 &2.71 &0.60 &0.61 &0.50 &1.54 &0.82 &0.53 &2.61 &1.93 &0.91 &4.55 &1.01 &1.07 &2.00 &1.47$ \pm $2.31 \\
   BASE-LaOML-$\sigma$10    &0.60 &0.92 &2.17 &1.56 &1.15 &2.72 &0.70 &0.64 &0.52 &1.58 &0.84 &0.52 &2.60 &1.93 &0.91 &4.58 &1.08 &1.07 &1.12 &1.43$ \pm $1.33 \\
   BASE-LaOML-$\sigma$15    &0.63 &0.98 &2.15 &1.65 &1.17 &2.74 &0.66 &0.66 &0.53 &1.50 &0.87 &0.59 &2.62 &1.97 &0.91 &4.62 &0.99 &1.16 &1.29 &1.46$ \pm $1.33 \\
   \hline
  \end{tabular}
\end{table*}

\subsubsection{The early-stop strategy} 
\begin{figure}[!htb]
 \centering
 \subfigure[]{%
 \begin{minipage}[!htb]{0.48\columnwidth}
 \centering
 \includegraphics[width=\columnwidth]{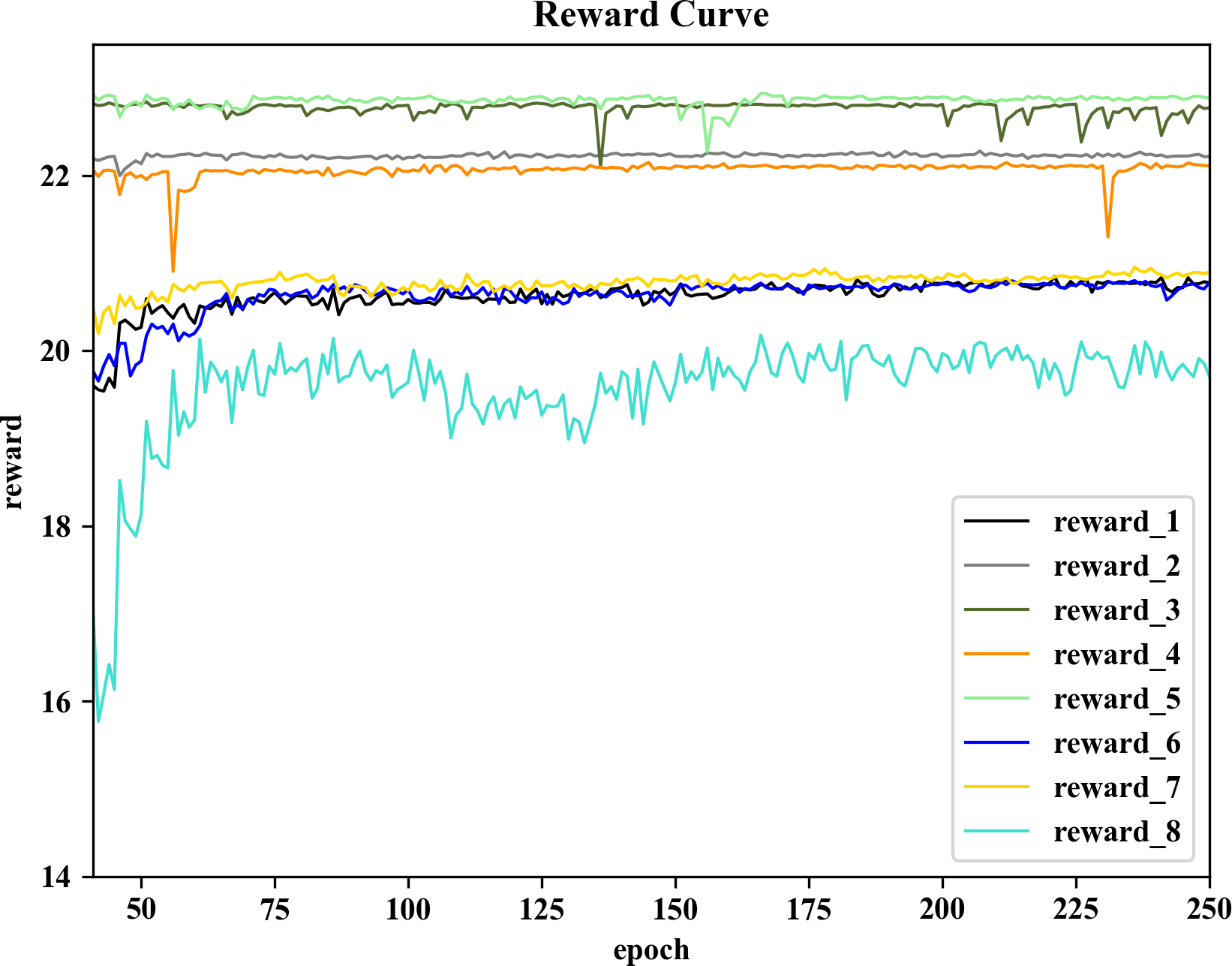}
 \end{minipage}
 }
 \subfigure[]{%
 \begin{minipage}[!htb]{0.48\columnwidth}
 \centering
 \includegraphics[width=\columnwidth]{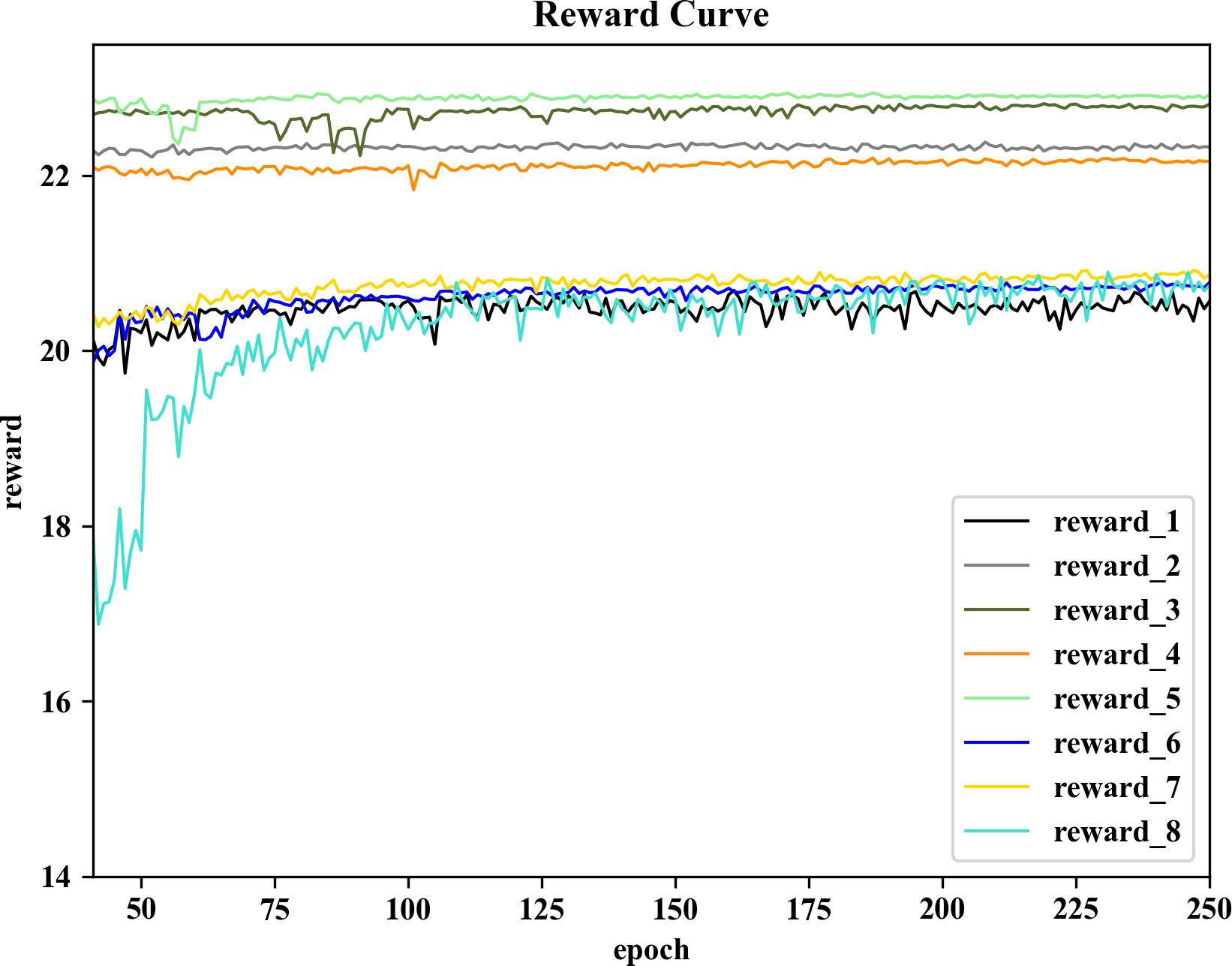}
 \end{minipage}
 }
 \caption{Comparison of reward without and with the early-stop strategy on the prenatal ultrasound dataset. (a) reward curves of the method without early-stop strategy; (b) reward curves of the method with early-stop strategy. The reward curve of the $n$th landmark is labeled as 'reward\_n'.}
 \label{reward_US}
\end{figure}

\begin{figure}[!htb]
 \centering
 \subfigure[]{%
 \begin{minipage}[!htb]{0.9\columnwidth}
 \centering
 \includegraphics[width=\columnwidth]{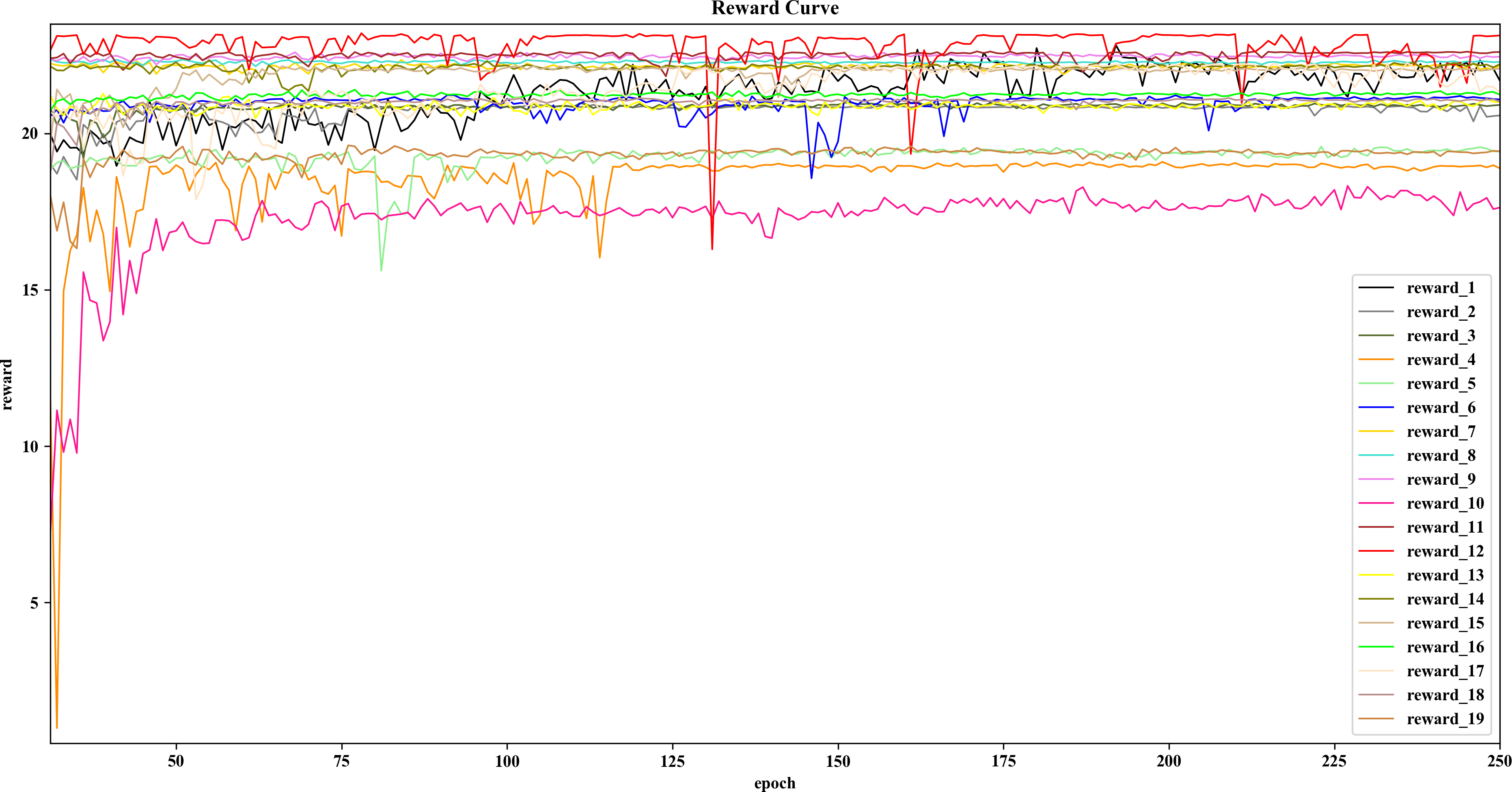}
 \end{minipage}
 }

 \subfigure[]{%
 \begin{minipage}[!htb]{\columnwidth}
 \centering
 \includegraphics[width=0.9\columnwidth]{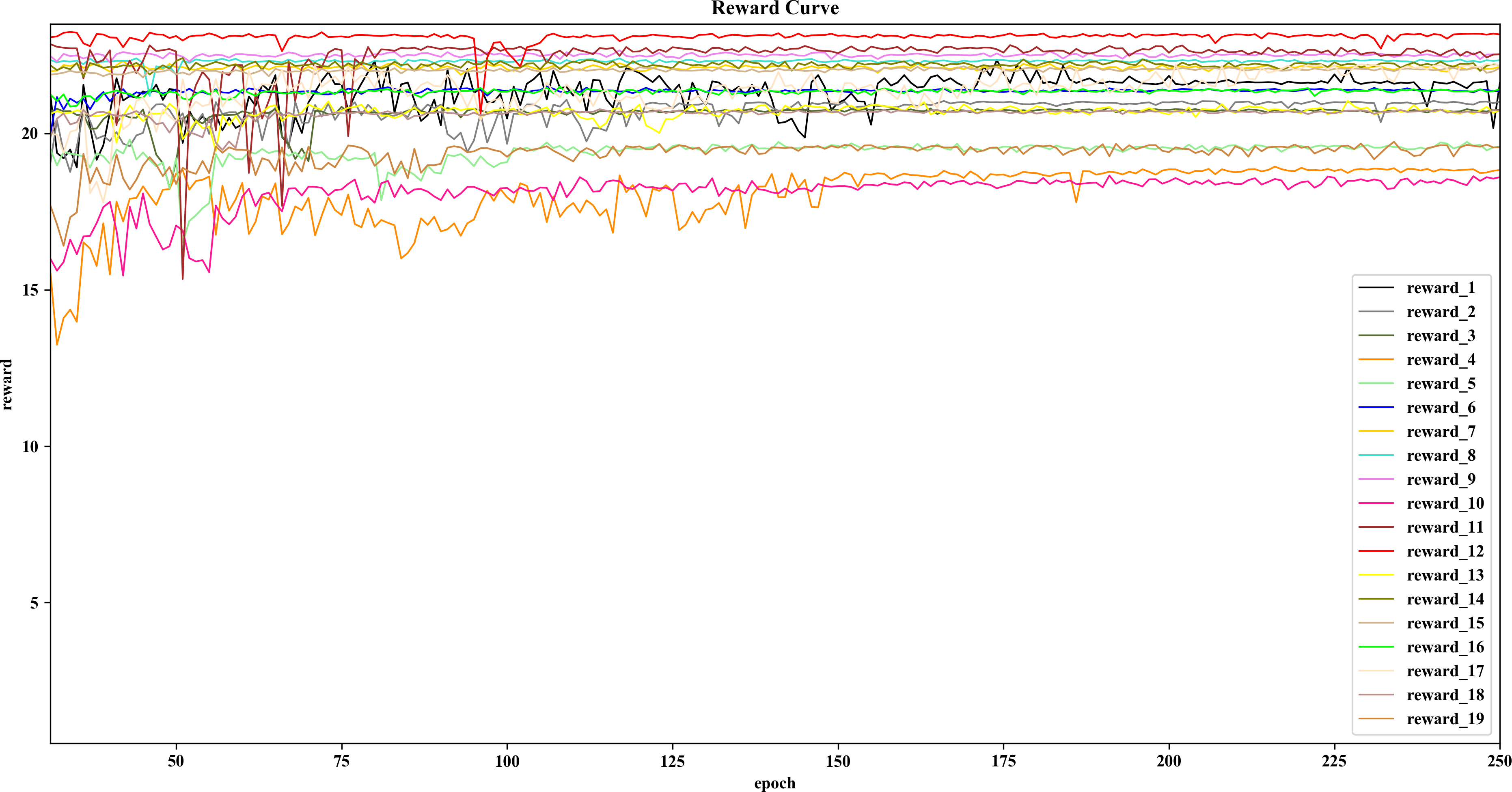}
 \end{minipage}
 }
 \caption{Comparison of reward between methods without and with early-stop strategy on the ACXRLDC dataset. (a) reward curves of the method without early-stop strategy; (b) reward curves of the method with early-stop strategy. The reward curve of $n$th landmark is labeled as 'reward\_n'.}
 \label{reward_X}
\end{figure}

The early-strop indicator is dependent on the variance of distance error. Fig. \ref{reward_US} and Fig. \ref{reward_X} presents the reward curve with and without the early-stop strategy. These results demonstrate that the early-stop tends to a more stable reward with smaller variances. Note that even without the early-stop strategy, our RL-based method also results in the upward trend of the reward curve, implying the effectiveness of our LaOML method.

%

\section{Discussion}
Our proposed learning-to-learn framework provides a novel and fundamental heatmap regression model to optimize the objective metrics for landmark localization in medical images with an RL method. Most current works focus on improving architectures for acquiring high-resolution features and high-level semantic information, neglecting another essence of heatmap regression for landmark localization. In this framework, we devise an RL-based objective metrics learning method with an early-stop strategy to take the localization accuracy as weakly supervision to tune variances of each landmark as the hyperparameters, thus improving localization accuracy and efficiency. We have evaluated our proposed framework on two datasets that contain 2D radiographs and 2D ultrasound images of different anatomical structures, demonstrating the generalization of our LaOML method. The experimental results reveal that our method outperforms the state-of-the-art approaches.

The key in the proposed model is to incorporate our novel learnable heatmap into heatmap regression in a unified framework. Improvements in localization accuracy mainly come from our RL-based LaOML method. By concurrently optimizing landmark localization and refining target precision, we obviate to make the tradeoff between a blurry and dense response and an accurate but highly sparse heatmap. This method is different from the learnable heatmap of the regularization \cite{payer2019integrating} and Adaloss \cite{teixeira2019adaloss}, where each of their networks manipulates the objective function to adjust the $\sigma$ for each landmark. However, the Adaloss only gradually decreases $\sigma$ for each target independently without considering the impact of $\Delta \sigma $ on MSE loss. It can also make the training difficult when dramatically changing in MSE loss variances. By contrast, Payer et al. \cite{payer2019integrating} optimize $\sigma$ and network parameters together using a unified objective function. The regularization term in the objective function tends to smaller $\sigma$, leading to weight sparsity and making the model hard to train. The unified objective function also introduces other hyperparameters to unite the landmark localization and heatmap refinement implicitly, increasing the difficulty of getting proper localization heatmap. Previous literature \cite{chen2019cephalometric, payer2019integrating} reveals that the regularization method only achieves limited improvement compared to the methods with fixed $\sigma$ \cite{arik2017fully, ibragimov2015computerized, lindner2015fully}. On the other hand, our RL-based LaOML method can decouple the optimization of the landmark localization and target precision explicitly in an end-to-end framework. It is easy to train on the target task without tunning the additional hyperparameter for refining heatmap ground truth. Extensive experiments illustrate that our proposed method can achieve good localization performance with a better balance between heatmap regression and heatmap refinement.

The variance decrease in loss suggests that the current task is close to convergence, which would help to indicate the optimal predication. Inspired by using the loss variance as the task convergence factor, in our proposed early-stop strategy, the constraints on RL agents are automatically obtained from the MRE metric variance on the validation dataset and integrated inside the heatmap refinement. We can see that the early-stop strategy has impacted on our proposed LaOML method from the experiments. Previous approaches in medical image analysis \cite{dou2019agent} introduce the RNN to map between the Q-value sequence and optimal step in value-function-based deep Q-Network (DQN) \cite{mnih2015human}. Nevertheless, differently to the complicated DQN methods, we employ policy gradient methods \cite{williams1992simple} to the fundamental heatmap refinement problem, improving its accuracy and efficiency directly.

The experimental results demonstrate that our learning-to-learning framework outperforms other state-of-the-art approaches on the publicly available ACXRLDC datasets. Interestingly, the baseline model using U-Net for heatmap regression shows competitive localization performance in the experiments. We think that this is due to the robust encoding and decoding ability of the multi-scale U-Net to extract salient high-resolution structures. Moreover, our baseline U-Net has much better performance than the Localization U-Net in \cite{payer2019integrating}. The main difference between these two U-Net lies in the total channel number among all intermediate convolution layers. The Localization U-Net only uses 128 channel numbers, whereas we follow the state-of-the-art U-Net with the maximum 1024 channel number. A recent study suggests that the network width also plays a vital role in capturing more fine-grained features \cite{tan2019efficientnet}. In line with the previous report in \cite{payer2019integrating}, this finding demonstrates the hypothesis that a high-resolution feature map is critical to landmark detection when using heatmaps \cite{xiao2018simple}.

The RNN and GAN based refinement methods bring some improvements in MRE over the BASE. These two refinement methods are also robust towards landmark misidentification by exploring dependencies among feature maps of different landmarks. Nonetheless, our LaOML surpasses them in both MRE and PCK, especially regarding localization accuracy. A possible explanation is that the RNN-based method requires a more complicated and task-specific relation among landmarks, whereas the GAN-based method adopts the shape prior instead of localization constraints. On the other hand, the BASE-C method shows worse localization performance than heatmap regression methods in the experiments. We conjecture that this result can be attributed to the fact that the MSE loss used in BASE-C is only calculated at the final coordinates, causing a small gradient for each pixel on the image. By contrast, the heatmap regression method directly applies MSE on each pixel, even though introducing the Gaussian distributions' spiny standard deviation parameters. Moreover, like coordinate regression methods, the BASE-C method directly learns a map from the landmark's appearance to its image coordinates, increasing the training difficulty and leading to limited performance, which has been reported in previous literature \cite{teixeira2019adaloss, noothout2020deep, pfister2015flowing}.

We employ the same additive update for all landmarks, in which $\sigma$ is only updated based on its current value and the predefined exploration range. However, landmarks that are detected unambiguously prefer smaller steps than those that show more ambiguity in the training dataset. In this study, the fixed exploration range to tune $\sigma$ overlooks these landmark-specific differences, restricting the search space. Therefore, this additive update might result in the local minima of heatmap refinement, making our heatmap ground truth prediction dependent on the initial value. The landmark-specific update, similar to the dynamic learning rate, should be further investigated in future studies with better capturing prior anatomical knowledge. Moreover, MSE loss is insensitive to small standard deviation compared to large $\sigma$, which hurts the capability to locate landmarks correctly. A possible solution is to design a novel loss function to amplify the influence of small $\sigma$ to localization.

\section{Conclusion}
This paper proposes a general learning-to-learn framework to optimize the localization and target precision of multiple landmarks simultaneously in medical images with an RL method. We leverage the RL-based LaOML strategy to use the localization accuracy as weakly supervised to predict the target heatmap of each landmark. An early-stop strategy is also advanced for active termination of the RL agent's interaction to balance the exploration-exploitation. Extensive experiments on the in-house prenatal US dataset and the publicly available ACXRLDC dataset prove the feasibility and effectiveness of our proposed learning-to-learn framework. In the future, we will extend the proposed learning-to-learn framework to more general landmark-detection tasks. More effort will be involved to further improve the localization, especially in exploring the landmark-specific update strategy and novel loss functions.

\bibliographystyle{IEEEtran}
\bibliography{ref}

\begin{thebibliography}{10}
\providecommand{\url}[1]{#1}
\csname url@rmstyle\endcsname
\providecommand{\newblock}{\relax}
\providecommand{\bibinfo}[2]{#2}
\providecommand\BIBentrySTDinterwordspacing{\spaceskip=0pt\relax}
\providecommand\BIBentryALTinterwordstretchfactor{4}
\providecommand\BIBentryALTinterwordspacing{\spaceskip=\fontdimen2\font plus
\BIBentryALTinterwordstretchfactor\fontdimen3\font minus
  \fontdimen4\font\relax}
\providecommand\BIBforeignlanguage[2]{{%
\expandafter\ifx\csname l@#1\endcsname\relax
\typeout{** WARNING: IEEEtran.bst: No hyphenation pattern has been}%
\typeout{** loaded for the language `#1'. Using the pattern for}%
\typeout{** the default language instead.}%
\else
\language=\csname l@#1\endcsname
\fi
#2}}

\bibitem{shen2017deep}
D.~Shen, G.~Wu, and H.-I. Suk, ``Deep learning in medical image analysis,''
  \emph{Annual review of biomedical engineering}, vol.~19, pp. 221--248, 2017.

\bibitem{khan2020survey}
A.~Khan, A.~Sohail, U.~Zahoora, and A.~S. Qureshi, ``A survey of the recent
  architectures of deep convolutional neural networks,'' \emph{Artificial
  Intelligence Review}, pp. 1--62, 2020.

\bibitem{huang2020segmentation}
Q.~Huang, Y.~Huang, Y.~Luo, F.~Yuan, and X.~Li, ``Segmentation of breast
  ultrasound image with semantic classification of superpixels,'' \emph{Medical
  Image Analysis}, vol.~61, p. 101657, 2020.

\bibitem{yin2019domain}
X.~Yin, Q.~Zhao, J.~Liu, W.~Yang, J.~Yang, G.~Quan, Y.~Chen, H.~Shu, L.~Luo,
  and J.-L. Coatrieux, ``Domain progressive 3d residual convolution network to
  improve low-dose ct imaging,'' \emph{IEEE transactions on medical imaging},
  vol.~38, no.~12, pp. 2903--2913, 2019.

\bibitem{2016A}
C.~W. Wang, C.~T. Huang, J.~H. Lee, C.~H. Li, S.~W. Chang, M.~J. Siao, T.~M.
  Lai, B.~Ibragimov, T.~Vrtovec, and O.~a. Ronneberger, ``A benchmark for
  comparison of dental radiography analysis algorithms,'' \emph{Medical Image
  Analysis}, vol.~31, pp. 63--76, 2016.

\bibitem{ghesu2017multi}
F.-C. Ghesu, B.~Georgescu, Y.~Zheng, S.~Grbic, A.~Maier, J.~Hornegger, and
  D.~Comaniciu, ``Multi-scale deep reinforcement learning for real-time
  3d-landmark detection in ct scans,'' \emph{IEEE transactions on pattern
  analysis and machine intelligence}, vol.~41, no.~1, pp. 176--189, 2017.

\bibitem{du2018articulated}
X.~Du, T.~Kurmann, P.-L. Chang, M.~Allan, S.~Ourselin, R.~Sznitman, J.~D.
  Kelly, and D.~Stoyanov, ``Articulated multi-instrument 2-d pose estimation
  using fully convolutional networks,'' \emph{IEEE transactions on medical
  imaging}, vol.~37, no.~5, pp. 1276--1287, 2018.

\bibitem{chen2019cephalometric}
R.~Chen, Y.~Ma, N.~Chen, D.~Lee, and W.~Wang, ``Cephalometric landmark
  detection by attentive feature pyramid fusion and regression-voting,'' in
  \emph{International Conference on Medical Image Computing and
  Computer-Assisted Intervention}.\hskip 1em plus 0.5em minus 0.4em\relax
  Springer, 2019, pp. 873--881.

\bibitem{zhou2020single}
G.-Q. Zhou, E.-Z. Huo, M.~Yuan, P.~Zhou, R.-L. Wang, K.-N. Wang, Y.~Chen, and
  X.-P. He, ``A single-shot region-adaptive network for myotendinous junction
  segmentation in muscular ultrasound images,'' \emph{IEEE Transactions on
  Ultrasonics, Ferroelectrics, and Frequency Control}, 2020.

\bibitem{zhang2017alzheimer}
J.~Zhang, M.~Liu, L.~An, Y.~Gao, and D.~Shen, ``Alzheimer's disease diagnosis
  using landmark-based features from longitudinal structural mr images,''
  \emph{IEEE journal of biomedical and health informatics}, vol.~21, no.~6, pp.
  1607--1616, 2017.

\bibitem{guo2020robust}
X.~Guo, H.~Wang, X.~Lu, X.~Hu, S.~Che, and Y.~Lu, ``Robust fovea localization
  based on symmetry measure,'' \emph{IEEE Journal of Biomedical and Health
  Informatics}, vol.~24, no.~8, pp. 2292--2302, 2020.

\bibitem{wu2017facial}
Y.~Wu, T.~Hassner, K.~Kim, G.~Medioni, and P.~Natarajan, ``Facial landmark
  detection with tweaked convolutional neural networks,'' \emph{IEEE
  transactions on pattern analysis and machine intelligence}, vol.~40, no.~12,
  pp. 3067--3074, 2017.

\bibitem{newell2016stacked}
A.~Newell, K.~Yang, and J.~Deng, ``Stacked hourglass networks for human pose
  estimation,'' in \emph{European conference on computer vision}.\hskip 1em
  plus 0.5em minus 0.4em\relax Springer, 2016, pp. 483--499.

\bibitem{teixeira2019adaloss}
B.~Teixeira, B.~Tamersoy, V.~Singh, and A.~Kapoor, ``Adaloss: Adaptive loss
  function for landmark localization,'' \emph{arXiv preprint arXiv:1908.01070},
  2019.

\bibitem{noothout2020deep}
J.~M. Noothout, B.~D. De~Vos, J.~M. Wolterink, E.~M. Postma, P.~A. Smeets,
  R.~A. Takx, T.~Leiner, M.~A. Viergever, and I.~I{\v{s}}gum, ``Deep
  learning-based regression and classification for automatic landmark
  localization in medical images,'' \emph{IEEE Transactions on Medical
  Imaging}, 2020.

\bibitem{pfister2015flowing}
T.~Pfister, J.~Charles, and A.~Zisserman, ``Flowing convnets for human pose
  estimation in videos,'' in \emph{Proceedings of the IEEE International
  Conference on Computer Vision}, 2015, pp. 1913--1921.

\bibitem{alansary2019evaluating}
A.~Alansary, O.~Oktay, Y.~Li, L.~Le~Folgoc, B.~Hou, G.~Vaillant, K.~Kamnitsas,
  A.~Vlontzos, B.~Glocker, B.~Kainz, \emph{et~al.}, ``Evaluating reinforcement
  learning agents for anatomical landmark detection,'' \emph{Medical image
  analysis}, vol.~53, pp. 156--164, 2019.

\bibitem{zhang2020enhanced}
M.~Zhang, J.~Xu, E.~A. Turk, P.~E. Grant, P.~Golland, and E.~Adalsteinsson,
  ``Enhanced detection of fetal pose in 3d mri by deep reinforcement learning
  with physical structure priors on anatomy,'' in \emph{International
  Conference on Medical Image Computing and Computer-Assisted
  Intervention}.\hskip 1em plus 0.5em minus 0.4em\relax Springer, 2020, pp.
  396--405.

\bibitem{xiao2018simple}
B.~Xiao, H.~Wu, and Y.~Wei, ``Simple baselines for human pose estimation and
  tracking,'' in \emph{Proceedings of the European conference on computer
  vision (ECCV)}, 2018, pp. 466--481.

\bibitem{arik2017fully}
S.~{\"O}. Arik, B.~Ibragimov, and L.~Xing, ``Fully automated quantitative
  cephalometry using convolutional neural networks,'' \emph{Journal of Medical
  Imaging}, vol.~4, no.~1, p. 014501, 2017.

\bibitem{chen2018cascaded}
Y.~Chen, Z.~Wang, Y.~Peng, Z.~Zhang, G.~Yu, and J.~Sun, ``Cascaded pyramid
  network for multi-person pose estimation,'' in \emph{Proceedings of the IEEE
  conference on computer vision and pattern recognition}, 2018, pp. 7103--7112.

\bibitem{sun2019deep}
K.~Sun, B.~Xiao, D.~Liu, and J.~Wang, ``Deep high-resolution representation
  learning for human pose estimation,'' in \emph{Proceedings of the IEEE
  conference on computer vision and pattern recognition}, 2019, pp. 5693--5703.

\bibitem{payer2016regressing}
C.~Payer, D.~{\v{S}}tern, H.~Bischof, and M.~Urschler, ``Regressing heatmaps
  for multiple landmark localization using cnns,'' in \emph{International
  Conference on Medical Image Computing and Computer-Assisted
  Intervention}.\hskip 1em plus 0.5em minus 0.4em\relax Springer, 2016, pp.
  230--238.

\bibitem{payer2019integrating}
------, ``Integrating spatial configuration into heatmap regression based cnns
  for landmark localization,'' \emph{Medical image analysis}, vol.~54, pp.
  207--219, 2019.

\bibitem{lampert2016empirical}
T.~A. Lampert, A.~Stumpf, and P.~Gan{\c{c}}arski, ``An empirical study into
  annotator agreement, ground truth estimation, and algorithm evaluation,''
  \emph{IEEE Transactions on Image Processing}, vol.~25, no.~6, pp. 2557--2572,
  2016.

\bibitem{zoph2018learning}
B.~Zoph, V.~Vasudevan, J.~Shlens, and Q.~V. Le, ``Learning transferable
  architectures for scalable image recognition,'' in \emph{Proceedings of the
  IEEE conference on computer vision and pattern recognition}, 2018, pp.
  8697--8710.

\bibitem{liu2018progressive}
C.~Liu, B.~Zoph, M.~Neumann, J.~Shlens, W.~Hua, L.-J. Li, L.~Fei-Fei,
  A.~Yuille, J.~Huang, and K.~Murphy, ``Progressive neural architecture
  search,'' in \emph{Proceedings of the European Conference on Computer Vision
  (ECCV)}, 2018, pp. 19--34.

\bibitem{ronneberger2015u}
O.~Ronneberger, P.~Fischer, and T.~Brox, ``U-net: Convolutional networks for
  biomedical image segmentation,'' in \emph{International Conference on Medical
  image computing and computer-assisted intervention}.\hskip 1em plus 0.5em
  minus 0.4em\relax Springer, 2015, pp. 234--241.

\bibitem{williams1992simple}
R.~J. Williams, ``Simple statistical gradient-following algorithms for
  connectionist reinforcement learning,'' \emph{Machine learning}, vol.~8, no.
  3-4, pp. 229--256, 1992.

\bibitem{li2019lfs}
C.~Li, X.~Yuan, C.~Lin, M.~Guo, W.~Wu, J.~Yan, and W.~Ouyang, ``Am-lfs: Automl
  for loss function search,'' in \emph{Proceedings of the IEEE International
  Conference on Computer Vision}, 2019, pp. 8410--8419.

\bibitem{xu2018less}
Z.~Xu, Y.~Huo, J.~Park, B.~Landman, A.~Milkowski, S.~Grbic, and S.~Zhou, ``Less
  is more: Simultaneous view classification and landmark detection for
  abdominal ultrasound images,'' in \emph{International Conference on Medical
  Image Computing and Computer-Assisted Intervention}.\hskip 1em plus 0.5em
  minus 0.4em\relax Springer, 2018, pp. 711--719.

\bibitem{chen2017adversarial}
Y.~Chen, C.~Shen, X.-S. Wei, L.~Liu, and J.~Yang, ``Adversarial posenet: A
  structure-aware convolutional network for human pose estimation,'' in
  \emph{Proceedings of the IEEE International Conference on Computer Vision},
  2017, pp. 1212--1221.

\bibitem{liu2019feature}
J.~Liu, H.~Ding, A.~Shahroudy, L.-Y. Duan, X.~Jiang, G.~Wang, and A.~C. Kot,
  ``Feature boosting network for 3d pose estimation,'' \emph{IEEE transactions
  on pattern analysis and machine intelligence}, vol.~42, no.~2, pp. 494--501,
  2019.

\bibitem{jakab2018unsupervised}
T.~Jakab, A.~Gupta, H.~Bilen, and A.~Vedaldi, ``Unsupervised learning of object
  landmarks through conditional image generation,'' \emph{arXiv preprint
  arXiv:1806.07823}, 2018.

\bibitem{song2020automatic}
Y.~Song, X.~Qiao, Y.~Iwamoto, and Y.-w. Chen, ``Automatic cephalometric
  landmark detection on x-ray images using a deep-learning method,''
  \emph{Applied Sciences}, vol.~10, no.~7, p. 2547, 2020.

\bibitem{ibragimov2015computerized}
B.~Ibragimov, B.~Likar, F.~Pernus, and T.~Vrtovec, ``Computerized cephalometry
  by game theory with shape-and appearance-based landmark refinement,'' in
  \emph{Proceedings of International Symposium on Biomedical imaging (ISBI)},
  2015.

\bibitem{lindner2015fully}
C.~Lindner and T.~F. Cootes, ``Fully automatic cephalometric evaluation using
  random forest regression-voting,'' in \emph{IEEE International Symposium on
  Biomedical Imaging}.\hskip 1em plus 0.5em minus 0.4em\relax Citeseer, 2015.

\bibitem{zoph2016neural}
B.~Zoph and Q.~V. Le, ``Neural architecture search with reinforcement
  learning,'' \emph{arXiv preprint arXiv:1611.01578}, 2016.

\bibitem{baker2016designing}
B.~Baker, O.~Gupta, N.~Naik, and R.~Raskar, ``Designing neural network
  architectures using reinforcement learning,'' \emph{arXiv preprint
  arXiv:1611.02167}, 2016.

\bibitem{dou2019agent}
H.~Dou, X.~Yang, J.~Qian, W.~Xue, H.~Qin, X.~Wang, L.~Yu, S.~Wang, Y.~Xiong,
  P.-A. Heng, \emph{et~al.}, ``Agent with warm start and active termination for
  plane localization in 3d ultrasound,'' in \emph{International Conference on
  Medical Image Computing and Computer-Assisted Intervention}.\hskip 1em plus
  0.5em minus 0.4em\relax Springer, 2019, pp. 290--298.

\bibitem{mnih2015human}
V.~Mnih, K.~Kavukcuoglu, D.~Silver, A.~A. Rusu, J.~Veness, M.~G. Bellemare,
  A.~Graves, M.~Riedmiller, A.~K. Fidjeland, G.~Ostrovski, \emph{et~al.},
  ``Human-level control through deep reinforcement learning,'' \emph{nature},
  vol. 518, no. 7540, pp. 529--533, 2015.

\bibitem{tan2019efficientnet}
M.~Tan and Q.~V. Le, ``Efficientnet: Rethinking model scaling for convolutional
  neural networks,'' \emph{arXiv preprint arXiv:1905.11946}, 2019.

\end{thebibliography}

\end{document}